\pdfoutput=1
\documentclass[11pt]{article}
\usepackage[preprint]{acl}
\makeatletter\ifacl@anonymize\else\fi\makeatother

\usepackage{times}
\usepackage{latexsym}
\usepackage[T1]{fontenc}
\usepackage[utf8]{inputenc}
\usepackage{microtype}

\usepackage{graphicx}
\usepackage{booktabs}
\usepackage{multirow}
\usepackage{array}
\usepackage{longtable}
\usepackage{multicol}
\usepackage{amsmath}
\usepackage{amssymb}
\usepackage{amsthm}
\usepackage{xcolor}
\usepackage{colortbl}
\definecolor{hlpass}{HTML}{D9F0F2}%
\definecolor{hlfail}{HTML}{FBE0DA}%
\definecolor{hlkey}{HTML}{FFF0B8}%
\definecolor{hlctrl}{HTML}{ECECEC}%
\definecolor{hlhead}{HTML}{E6EFF9}%
\newcommand{\hlpass}{\cellcolor{hlpass}}
\newcommand{\hlfail}{\cellcolor{hlfail}}
\newcommand{\hlkey}{\cellcolor{hlkey}}

\newcommand{\hlheadrow}{\rowcolor{hlhead}}

\newcommand{\hlctrlrow}{\rowcolor{hlctrl}}
\definecolor{hlgroup}{HTML}{ECE8F7}%
\newcommand{\hlgrouprow}{\rowcolor{hlgroup}}
\newcommand{\hlbest}[1]{{\boldmath\textbf{#1}}}%
\newcommand{\hlhd}[1]{{\boldmath\textbf{#1}}}%
\ExplSyntaxOn
\box_if_exist:NF \l__hlclip_box { \box_new:N \l__hlclip_box }
\DeclareDocumentCommand \hlclip { +m }
  {
    \hbox_set:Nn \l__hlclip_box { #1 \unskip }
    \box_set_clipped:N \l__hlclip_box
    \mode_leave_vertical:
    \box_use_drop:N \l__hlclip_box
  }
\ExplSyntaxOff

\usepackage{url}
\usepackage{xspace}
\usepackage{enumitem}
\usepackage[capitalize,noabbrev]{cleveref}

\theoremstyle{remark}

\newcounter{remark}
\newcommand{\remarkpar}[1]{\par\noindent\refstepcounter{remark}\textit{Remark~\theremark.}\label{#1}\ }

\newcommand{\sk}[1]{\texttt{[#1]}}

\title{When Skills Fail to Compose:\\Compatibility and Retention in Language Models}

\author{%
  \textbf{Yan Lin}\textsuperscript{\textnormal{1,2,*}}\quad
  \textbf{Zhongliang Guo}\textsuperscript{\textnormal{3,*}}\quad
  \textbf{Jingyu Sun}\textsuperscript{\textnormal{4,5}}\quad
  \textbf{Qiyan Zhao}\textsuperscript{\textnormal{6}}\\
  \textbf{Ruihong Zeng}\textsuperscript{\textnormal{7}}\quad
  \textbf{Ying Chen}\textsuperscript{\textnormal{8}}\quad
  \textbf{Zhengtao Yao}\textsuperscript{\textnormal{9}}\quad
  \textbf{Yan Wen}\textsuperscript{\textnormal{9}}\\
  \textbf{Kunyu Peng}\textsuperscript{\textnormal{10}}\quad
  \textbf{Yuxia Wang}\textsuperscript{\textnormal{1,\textdagger}}\\[0.5mm]
  \textsuperscript{1} INSAIT, Sofia University ``St. Kliment Ohridski''\quad
  \textsuperscript{2} Newcastle University\\
  \textsuperscript{3} University of Aberdeen\quad
  \textsuperscript{4} University of Manchester\quad
  \textsuperscript{5} University of Melbourne\\
  \textsuperscript{6} Chinese Academy of Sciences\quad
  \textsuperscript{7} MBZUAI\quad
  \textsuperscript{8} Arizona State University\\
  \textsuperscript{9} University of Southern California\quad
  \textsuperscript{10} Karlsruhe Institute of Technology\\
  \texttt{\{y.lin64@ncl.ac.uk, yuxia.wang@insait.ai\}}%
}

\begin{document}
\maketitle

\makeatletter
\ifacl@anonymize\else
\begingroup
\renewcommand{\thefootnote}{\fnsymbol{footnote}}
\footnotetext[1]{Equal contribution.}
\footnotetext[2]{Corresponding author.}
\endgroup
\fi
\makeatother

\begin{abstract}
Skill acquisition alone does not establish a reusable skill library: task proficiency does not specify a compositional interface. Motivated by agents that acquire and refine skills, we study independently learned soft-prompt skills on frozen language models, with skill identities and execution order supplied. Across four digit-task bases and three record-editing models, mean composition accuracy in one call is at most 7.6\%, compared with 88.2\% to 100\% over two calls that add computation and an explicit intermediate output. Neutral-insertion training improves robustness but leaves substantial interference, which block length and per-token norm alone cannot explain on two of four digit bases. Blocks of the same operation need not be interchangeable: after training next to frozen learned partners, a post hoc test on two Qwen bases shows that replacing those partners with independently learned blocks of the same operations lowers median composition accuracy from 98.0\% and 95.3\% to 12.7\% and 0.5\%, although every replacement scores at least 90\% alone. Repair can improve a composition while impairing the original skill, and rehearsal improves retention while transfer remains limited on Qwen2.5-0.5B. Internal interventions constrain explanations without identifying a shared failure mechanism. These findings motivate evaluating compatibility, retention and transfer alongside acquisition when building skill libraries for self-evolving agents. Code is available at \url{https://github.com/YanLin-Quinne/SkillCompatibility}.
\end{abstract}

\section{Introduction}
\label{sec:intro}
\begin{figure*}[t]
\centering
\includegraphics[width=\textwidth]{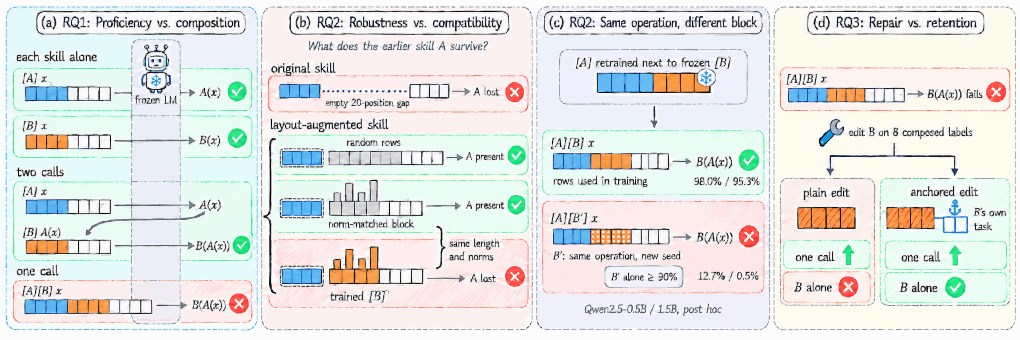}
\caption{\textbf{Task proficiency does not specify a compositional interface.} (a)~Skills that work alone and over two calls fail in one call. (b)~Layout alone can suppress $A$, and on two Qwen bases layout-augmented skills keep $A$ after a norm-matched block but not after trained $B$. (c)~A skill retrained next to frozen $B$ fails with $B'$, a proficient block of the same operation (post hoc, medians on 0.5B and 1.5B). (d)~Both edits raise one-call accuracy, but only the anchored edit, with twice the examples, keeps the edited skill.}
\label{fig:overview}
\vspace{-5pt}
\end{figure*}

Language models can acquire new skills as small learned parameter blocks, such as soft prompts, prefixes, new token embeddings and adapters, while the base model stays frozen~\citep{lester2021power,li2021prefix,hewitt2025neologism}. Recent agent systems store such blocks as reusable skills, including SkillSmith~\citep{dery2026skillsmith} and LatentSkill~\citep{yu2026latentskill}, so that a library of them can be recombined and revised as an agent's needs change. Realising this promise requires knowing which skills work together and which earlier behaviours survive revision, and successful acquisition alone establishes neither.

The explanatory challenge is that separately learned skills still act through a shared model. Their contributions may change when other skills enter the same context or when a skill is replaced by another learned implementation. Understanding reuse therefore requires examining the relationship between skill parameters, execution context and induced behaviour. We study these dependencies in a deliberately controlled setting: soft prompts, small blocks of learned input embeddings on a frozen language model~\citep{lester2021power}, trained on synthetic digit transformations and record edits whose intermediate results can be checked. The setting trades realism for control, so it tests model-level requirements that a library of learned blocks inherits rather than agent behaviour. One thesis organises the results: task proficiency does not specify a compositional interface. Two blocks that each perform the same operation alone need not be interchangeable in composition, and compatibility may depend on the particular learned instance and on the skills it runs with.

Skill Neologisms reports zero-shot composition of independently trained skills~\citep{berthon2026skill}. Our replication falls within its reported order-pooled range. Tailor identifies position-sensitive prompt interference and introduces a shared connector~\citep{yang2023tailor}. Building on these findings, we evaluate 27 non-degenerate directed digit-skill pairs with each order scored separately, then examine compatibility across skill implementations and after repair. Experiments cover four digit-task bases and three record-editing models, with skill identities and execution order supplied. Figure~\ref{fig:overview} organises three questions.

\textbf{RQ1: Do individually proficient skills compose reliably in one model call?} Mean composition accuracy is at most 7.6\%, compared with 88.2\% to 100\% over two calls (Section~\ref{sec:revisit}). The latter add computation and an explicit intermediate output. Restricting evaluation to proficient constituent skills preserves this gap. In a post hoc check, digit skills nevertheless compose accurately with pretrained operations encountered during training, indicating that failure depends on which skills are combined.

\textbf{RQ2: How does compatibility depend on context and skill implementation?} Neutral-insertion training improves robustness but does not establish compatibility. For layout-augmented skills, matched-block controls show that length and per-token norm alone do not explain the residual interference on two of four digit bases, with mixed results on the others (Section~\ref{sec:diag}). Training with other learned skills improves composition mainly with the blocks used during training. In a post hoc replacement test on two Qwen bases, replacing these blocks with independently learned implementations of the same operations lowers median composition accuracy from 98.0\% and 95.3\% to 12.7\% and 0.5\%, although every replacement scores at least 90\% alone. Separately, activation replacement and knockout experiments test candidate internal explanations for composition failure without establishing a shared failure mechanism.

\textbf{RQ3: Does repairing composition preserve reusable skills?} Repair can impair the original skill, while rehearsal with eight additional original-task examples per pair, doubling the plain edit's examples, improves retention without a consistent composition advantage (Section~\ref{sec:repair}). With more labels, anchored successor repair supports accurate compositions that retain both skills on seed 1 of the three smaller digit bases. However, for one set of independently learned skills on Qwen2.5-0.5B, repair with rehearsal using 256 composed labels per training pair reaches mean accuracy of 90.4\% on training combinations and 10.4\% with held-out predecessors. Successful repair can remain specific to the combinations used during training.

Our contribution is a controlled diagnosis of the dependencies that limit skill reuse: standalone proficiency, insertion robustness, success with training partners and repaired-pair accuracy each leave unspecified how a skill composes. These model-level findings motivate evaluating whether skill acquisition and revision expand reliably reusable capabilities in future self-evolving agents while preserving the behaviours on which existing uses depend.

\section{Related Work}
\label{sec:related}

\paragraph{Parametric skills on frozen models.}
Soft prompts and prefixes adapt frozen models with learned input vectors~\citep{lester2021power,li2021prefix}, and neologisms make them vocabulary entries for behaviours and skills~\citep{hewitt2025neologism,berthon2026skill}. Concept tokens learn behaviours from definitions and TokMem procedures from multistep data~\citep{sastre2026concept,wu2026tokmem}, and SkillSmith and LatentSkill build composable agent skills as prefixes and LoRA adapters~\citep{dery2026skillsmith,yu2026latentskill}. Adapters, LoRA modules and learned prompts are fused, routed, transferred or mixed~\citep{pfeiffer2021adapterfusion,zhang2023composing,huang2024lorahub,wu2024mole,vu2022spot,asai2022attempt,wang2026skim}. We test whether independently acquired skills stay compatible in one call.

\paragraph{Interference between learned prompts.}
Independently trained prefixes and attribute prompts can interfere~\citep{gu2022distributional,huang2023extensible}. Tailor reports position-dependent dominance and counters it with masking, position reindexing and a shared connector~\citep{yang2023tailor}, and single soft prompts are also position-sensitive~\citep{yang2025position}. Steering Tokens learns a shared token to compose behaviour tokens~\citep{radevski2026steering}. Skill Neologisms composes independently trained skills zero-shot, and Reasoning with Neologisms finds limits under chain-of-thought prompting~\citep{berthon2026skill,berthon2026reasoning}. We add order-resolved pairs, matched blocks, retention tests after repair and connectors shared at three levels.

\paragraph{Agent skill libraries.}
Agents acquire skills as code, API calls or text workflows~\citep{wang2024voyager,lu2025axis,wang2025awm,wang2025asi,zheng2025skillweaver}. The Agent Skills format and SkillsBench load several textual skills into one context~\citep{zhang2025agentskills,li2026skillsbench}, and Graph-of-Skills retrieves skill bundles from libraries of thousands~\citep{liu2026graphofskills}. Optimisers rewrite skills under validation gates~\citep{yang2026skillopt,moll2026grasp,chen2026skillspec}. Co-installed skills conflict, and bank-level gains can hide harmful combinations~\citep{yan2026oneskill,zhao2026coalition}. Self-evolution can erode earlier capabilities~\citep{yu2026selfevolving} and trade security for utility~\citep{li2026auditing}. SkillScriptBench benchmarks repair and preservation of executable skill packages, and composed text modules interfere in deployed agents~\citep{liu2026skillscriptbench,lin2026instruction}. We fix skill identities and order, removing selection errors, and study parametric skills rather than deployed text or code libraries.

\paragraph{Composition and internal computation.}
Component accuracy need not ensure composed accuracy~\citep{press2023measuring}, and prompted skills or stacked instructions degrade when combined in one generation~\citep{yu2024skillmix,anand2026stacking}. Explicit intermediate outputs and decomposed prompting can help~\citep{ramesh2024compositional,khot2023decomposed}. Function vectors carry composable task information~\citep{todd2024function}, and models compute two-hop functions through or around an intermediate value~\citep{khandelwal2025compose}. Activation patching depends on intervention design and metrics~\citep{zhang2024best}, so we do not equate restored behaviour with an identified pathway.

\section{Skills and Evaluation}
\label{sec:setup}

\paragraph{Ordered composition.}
Each skill is a separately learned block of 20 input embedding rows on a frozen model $\theta$, and the protocol supplies skill identities and execution order. Let $T_A$ denote skill $A$'s transformation and $f_\theta([A],x)$ the parsed output when its block precedes input $x$. In a directed pair $A\!\to\!B$, predecessor $A$ is applied first and successor $B$ second, each the other's partner, and the target is $T_B(T_A(x))$ even for non-commuting transformations. Over evaluation inputs, we measure
\begin{align}
P_A &= \Pr[f_\theta([A],x)=T_A(x)], \\
C_{A\to B} &= \Pr[f_\theta([A][B],x)=T_B(T_A(x))].
\end{align}
$P_A$ and $P_B$ measure standalone proficiency, and compatibility compares $C_{A\to B}$ with individual and two-call execution. Insertion robustness asks whether $A$'s effect survives an intervening block. After repair, retention remeasures the edited skill alone and transfer tests unseen partners.

\paragraph{Digit transformations.}
Six skills implement left and right shifts, adjacent swaps, modulo three, digit mirroring and inverted polarity. Following Skill Neologisms~\citep{berthon2026skill}, each is trained next to pretrained operations, but never another new skill. We evaluate the released Qwen2.5-0.5B digit base and Qwen2.5-1.5B, Qwen2.5-7B and Gemma-3-1B bases further pretrained on the same operations~\citep{qwen2024qwen25,gemma2025gemma3}. Excluding identity compositions and pairs with no observable second step leaves 27 directed pairs (18 commuting) and three skill seeds per base.

For example, \sk{SHIFT\_LEFT} maps 26647378 to 66473782, and applying \sk{SWAP\_PAIRS} next gives 66747328. Behavioural comparisons use 300 length-8 inputs per condition. The 1.5B and 7B digit bases kept early checkpoints after numerical divergence (Appendix~\ref{sec:app-setup}). We retain 7B seed 1's failed \sk{MOD3} in the full matrix and emphasise seeds 2 and 3 in later diagnostics and repair.

\paragraph{Record editing.}
Six skills edit a record's name or total on Qwen2.5-Instruct at 0.5B, 1.5B and 7B. Their 30 directed pairs comprise 18 different-field pairs, six commuting name pairs and six order-dependent total pairs. A pair is eligible when both skills reach 90\% alone: this retains 20 pairs at each smaller size and all 30 at 7B. Unlike digit skills, they are trained next to the identity filler only. Each skill also has a document version with a procedure and two worked examples, and the prompt loads the invoked documents and gives their order.

\paragraph{Outcomes and comparisons.}
All accuracies use exact match and greedy decoding, and digits are scored on the first gold-length output span. Two-call execution applies $B$ to the parsed output of $A$, testing sequential executability with an explicit intermediate representation. It is not matched to one call: digit prompts have 49 tokens in one call and 58 across two, and batched two-call evaluation takes 1.61 to 2.06 times as long (Appendix~\ref{sec:app-cost}). For digits, Table~\ref{tab:phenomenon} compares both modes on the same diagnostic subset, whose candidate outputs are pairwise distinct so that an omitted operation cannot receive credit by coincidence.

We also measure whether a constituent's \emph{effect is present in the output}. For digits, ``$A$ present'' accepts $A(x)$ or the known successor applied to $A(x)$. For records, a field-level flag marks the edit. These measures do not establish that an internal representation of $A(x)$ exists. Novel skill-token IDs are masked during decoding, since tied embeddings could otherwise emit them. On untied Qwen2.5-7B, masking changes no output (Appendix~\ref{sec:app-mask}).

\paragraph{Statistical scope.}
Skill seeds vary only the sampled training examples, not the initialisation (Appendix~\ref{sec:app-setup}). A digit pair interferes when one-call accuracy falls significantly below $P_AP_B$ in at least two seeds. The digit test ignores dependence between single and composed runs on shared inputs, so its $p$ values are approximate. An eligible record-editing pair interferes when one-call accuracy is at least 10 points below two calls and significantly lower in at least two seeds. Tests use Holm correction~\citep{holm1979simple}. Shared constituents mean pooled counts and pair tests are not independent replications. Logged rules were recorded internally before the runs, amended rules changed after earlier results, exploratory analyses fixed selection rules after initial runs, and post hoc analyses came later.

\section{Proficiency Is Not Compatibility}
\label{sec:revisit}
\textbf{RQ1} asks whether individually proficient skills compose reliably in one model call.
\begin{table}[t]
\centering
\small
\setlength{\tabcolsep}{4pt}
\hlclip{\begin{tabular}{@{}lrrr@{}}
\toprule
\hlheadrow \hlhd{Model} & \hlhd{One call} & \hlhd{Two calls} & \hlhd{Interfering} \\
\midrule
\hlgrouprow \multicolumn{4}{@{}l}{\emph{Digit transformations, matched diagnostic items}} \\
Qwen2.5-0.5B & 6.4 & \hlbest{100.0} & \hlfail 27/27 \\
Qwen2.5-1.5B & 1.3 & \hlbest{99.9} & \hlfail 27/27 \\
Qwen2.5-7B$^{\dagger}$ & 0.9 & \hlbest{88.2} & \hlfail 27/27 \\
Gemma-3-1B & 2.7 & \hlbest{100.0} & \hlfail 27/27 \\
\midrule
\hlgrouprow \multicolumn{4}{@{}l}{\emph{Record editing, eligible pairs}} \\
Qwen2.5-0.5B & 7.6 & \hlbest{100.0} & \hlfail 20/20 \\
Qwen2.5-1.5B & 1.1 & \hlbest{100.0} & \hlfail 20/20 \\
Qwen2.5-7B & 2.0 & \hlbest{99.0} & \hlfail 30/30 \\
\bottomrule
\end{tabular}}

\caption{\textbf{Two calls succeed where one call fails.} Means (\%) over directed pairs and three skill seeds, digits on the diagnostic subset. Counts are pairs meeting the interference criterion. $\dagger$:~includes failed 7B seed 1.}
\label{tab:phenomenon}
\vspace{-5pt}
\end{table}

\begin{figure*}[t]
\centering
\includegraphics[width=\textwidth]{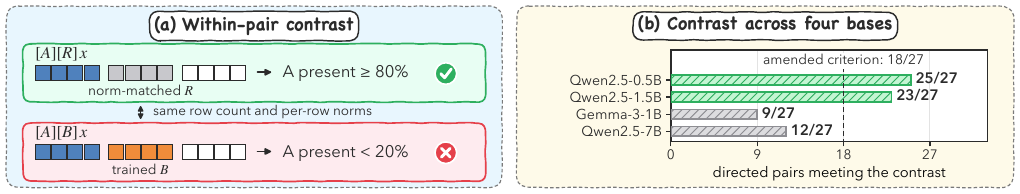}
\caption{\textbf{Insertion robustness does not certify compatibility on two digit bases.} Pairs where a norm-matched block keeps $A$ but trained $B$ removes it in most seeds, 7B seeds 2 and 3 (Appendix~\ref{sec:app-compat}).}
\label{fig:fig_compat}
\vspace{-5pt}
\end{figure*}

\subsection{Two Calls Succeed Where One Fails}
\label{sec:revisit-interfere}

One-call composition is unreliable across the tested skill libraries (Table~\ref{tab:phenomenon}). On matched inputs, mean one-call accuracy is 0.9\% to 6.4\% on the digit diagnostic subset and 1.1\% to 7.6\% on eligible record-editing pairs, against 88.2\% to 100\% and 99.0\% to 100\% over two calls. All digit pairs and eligible record-editing pairs meet their interference criteria.

Weak constituents do not explain the overall gap. Excluding the failed 7B seed raises 7B two-call accuracy to 98.1\%, and skills passing the revised acceptance rule still interfere significantly in all but one retained pair and seed cell (Appendix~\ref{sec:app-acceptance-scope}).

Familiar partners give a positive control. Post hoc, digit skills compose with pretrained operations met in training, with median one-call accuracy of 100\% in both orders on every base, although 26 of 108 skill, seed and operation cells on 1.5B fall below 90\% with the pretrained operation first (Appendix~\ref{sec:app-acceptance-scope}). This control changes the partner block and its familiarity and cannot separate them.

Pooled success need not mean per-order reliability. Under the Skill Neologisms protocol and base, our skills reach 12.5\% to 13.6\% in one order, 44.0\% to 89.8\% in the other (Appendix~\ref{sec:app-sn}).

\subsection{Errors Do Not Reveal a Shared Cause}
\label{sec:revisit-which}

Digit failures do not reduce to clean omission. Outputs match $B(x)$ more often than $A(x)$ (8.5\% to 12.0\% against 0.8\% to 4.6\%, chance at most 0.1\%), but most match neither. A post hoc graded comparison favours earlier-skill loss in only 16 to 18 of 27 pairs per base (sign-test $p$ 0.12 to 0.44, Appendix~\ref{sec:app-failure}).

Record-editing outputs more often retain the earlier edit: its presence rate is 60.3\% to 71.5\%, compared with 10.1\% to 30.2\% for the later edit. This reverses the logged earlier-loss prediction. Because task, base model and training neighbours change together, the contrast describes the outputs without establishing a domain-dependent~\mbox{mechanism}.

\noindent\textbf{Finding 1.} Standalone proficiency and accurate sequential execution do not certify one-call compatibility on any tested base, although digit skills compose with pretrained partners met in training, and errors identify no common failure~mechanism.

\section{Diagnosing Skill Incompatibility}
\label{sec:diag}

\textbf{RQ2} asks how compatibility depends on context and skill implementation. We vary layout, partner rows and representation, then test interventions.

\subsection{Layout Alone Can Suppress a Skill}
\label{sec:diag-layout}

A trained successor is not needed to make a proficient digit skill fail. We insert random rows, initial skill rows, identity tokens or neutral text between $A$ and its input. An empty position-ID gap varies separation without inserting content (Appendix~\ref{sec:app-d1}).

All 66 tested skill seeds keep $A$ present at 80\% or more alone, yet an empty 20-position gap lowers it below 20\% in 63 of them, with the three exceptions on Gemma-3-1B. Inserted material also suppresses many skills, although the effect varies with skill and span. On 0.5B seed 1, the gap also removes $A$ from every five- and 40-row skill (Appendix~\ref{sec:app-ll}).

These failures establish \emph{layout sensitivity}. The empty position-ID gap is absent from ordinary concatenation, so it cannot quantify the contribution of position changes to natural composition failure.

\subsection{Robust Skills Still Interfere}
\label{sec:diag-compat}

To test whether untrained insertions alone explain the failures, we retrain each digit skill independently with random rows or neutral text inserted at a random operation boundary before the digits in half the inputs. These \emph{layout-augmented} skills never encounter another newly trained skill.

The augmentation substantially improves insertion robustness. On Qwen2.5-0.5B, all 18 trained skill seeds preserve $A$ with an $A$-present rate of at least 80\% after 20 random rows, initial skill rows or copies of neutral text. Nevertheless, a trained successor suppresses $A$ below 20\% in 73 of 81 pair and seed cells. Every skill scores 100\% alone, while median one-call accuracy across pairs is 0.3\%. All four digit bases fail the logged composition criterion after augmentation (Appendix~\ref{sec:app-l1}).

\paragraph{Block length and norm alone do not explain the loss on two bases.}
To test whether the trained successor harms $A$ only through its length and scale, we compare it with a random block of the same row count and per-row norms. A directed pair shows the diagnostic contrast when most seeds preserve $A$ after this matched block ($\geq 80\%$) but lose it after the trained successor ($<20\%$).

Figure~\ref{fig:fig_compat} shows this contrast on 25 of 27 pairs on Qwen2.5-0.5B and 23 on Qwen2.5-1.5B. On both bases, $A$ present is significantly lower after the trained successor than after its matched block in every pair and seed (paired McNemar tests with Holm correction). Two more random directions on 0.5B give 25 and 24 such pairs.

On Gemma-3-1B and Qwen2.5-7B, only 9 and 12 pairs show it (7B seeds 2 and 3), below the amended threshold of 18, mostly because the matched block itself fails to preserve $A$. The contrast is thus bounded evidence about the tested blocks, not a universal separation or evidence that a particular learned direction causes the loss.

\subsection{Learned Blocks Need Not Be Interchangeable}
\label{sec:diag-partner}
\begin{figure*}[t]
\centering
\includegraphics[width=\textwidth]{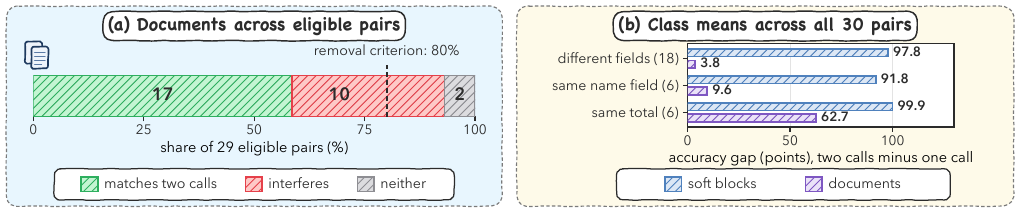}
\caption{\textbf{Documents help only some pairs.} (a)~7B pairs against the removal criterion of at least 80\% matching and no interfering pair. (b)~Post hoc 7B mean gaps (Appendix~\ref{sec:app-c1}).}
\label{fig:fig_carrier}
\vspace{-5pt}
\end{figure*}

\begin{table}[t]
\centering
\footnotesize
\setlength{\tabcolsep}{4pt}
\hlclip{\begin{tabular}{@{}lcccc@{}}
\toprule
\hlheadrow & \multicolumn{2}{c}{\hlhd{Seen operation}} & \multicolumn{2}{c}{\hlhd{Held-out operation}} \\
\cmidrule(lr){2-3}\cmidrule(l){4-5}
\hlheadrow \hlhd{Partner rows} & \hlhd{0.5B} & \hlhd{1.5B} & \hlhd{0.5B} & \hlhd{1.5B} \\
\midrule
Original & \hlpass 98.0 & \hlpass 95.3 & \hlfail 0.0 & \hlfail 0.0 \\
New$^\ast$ & \hlfail 12.7 & \hlfail 0.5 & \hlfail 0.0 & \hlfail 0.0 \\
\bottomrule
\end{tabular}
}
\caption{\textbf{Training next to learned partners largely does not carry over to new partner rows.} Medians (\%) over skill and partner seeds, then over 27 pairs. Every retrained skill and new row scores at least 90\% alone. $\ast$:~post hoc, rows of skill seeds 104 and 105.}
\label{tab:l2-body}
\vspace{-5pt}
\end{table}

We next ask whether compatibility gained with partners extends beyond the blocks met in training. On two Qwen bases, we retrain each skill next to frozen learned partners, holding some operations out. Median composition reaches 98.0\% and 95.3\% with the partner rows used in training but 0\% with held-out operations (Table~\ref{tab:l2-body}).

A post hoc test replaces the partner rows of seen operations with independently learned rows of the same operations from two new skill seeds. Every retrained skill and replacement scores at least 90\% alone, yet median composition falls to 12.7\% on 0.5B and 0.5\% on 1.5B (three skill seeds and two partner seeds under one recipe). This is consistent with compatibility forming mainly between the specific blocks that met in training rather than with a partner's operation (Appendices~\ref{sec:app-neo-reverse} and~\ref{sec:app-l2}). A learned \sk{REVERSE} block is an exception: it also composes with \sk{ADD}, unseen in training.

\subsection{Documents Help Some Skill Combinations}
\label{sec:diag-carrier}

Documents give the same transformations as procedures and worked examples, changing both representation and the information given at inference. Across eligible pairs, documents only partly close the record-editing gap, and no base meets the logged removal criterion. At 0.5B and 1.5B, they still interfere on four of four and nine of 12 pairs whose documents pass the single-skill gate. At 7B, they match two calls within 10 points on 17 of 29 pairs, one of them with zero accuracy in both modes, and ten pairs still interfere, including all six order-dependent pairs (Figure~\ref{fig:fig_carrier}).

A post hoc split suggests where documents help: on the 18 different-field 7B pairs, they match two calls on 16 while soft blocks interfere on all 18 (mean gaps 3.8 and 97.8 points). Loading all six documents, dropping worked examples or keeping only examples with skill names weakens this advantage. These controls implicate supplied task information but do not isolate an effect of text representation (Appendices~\ref{sec:app-document-examples} and~\ref{sec:app-c2}).

\subsection{Interventions Constrain Explanations}
\label{sec:diag-probes}
\begin{figure*}[t]
\centering
\includegraphics[width=\textwidth]{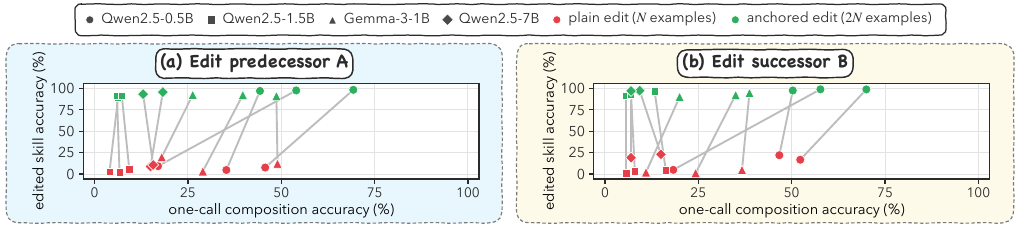}
\caption{\textbf{Repaired-pair accuracy does not certify retention.} Marginal medians at eight composed labels, not individual pairs, with lines joining plain and anchored edits of one base, seed and side. 7B seed 1 is excluded.}
\label{fig:fig_retention}
\vspace{-5pt}
\end{figure*}

\begin{table}[t]
\centering
\footnotesize
\setlength{\tabcolsep}{3pt}
\hlclip{\begin{tabular}{@{}ll@{}}
\toprule
\hlheadrow \hlhd{Test} & \hlhd{Result} \\
\midrule
\hlgrouprow \multicolumn{2}{@{}l}{\emph{Restoration, generation accuracy}} \\
Per-cell transplant$^\ast$ & \hlpass 19 of 30 cells gain $\geq 10$ points \\
\hlgrouprow \multicolumn{2}{@{}l}{\emph{Shared site, first-digit logit proxy}} \\
One site per base$^\dagger$ & \hlfail 7 of 39 cells beat every control \\
\hlgrouprow \multicolumn{2}{@{}l}{\emph{Necessity of the selected readouts}} \\
Mean ablation & \hlfail 32 of 32 cells unchanged \\
Steering & \hlfail no cell restored \\
\hlgrouprow \multicolumn{2}{@{}l}{\emph{Successor access and attention heads}} \\
Successor keys blocked & \hlfail $A$ lost in 22 of 22, as a gap \\
Early-layer key knockout & \hlfail 0 of 22 cells succeed \\
Few skill-reading heads & \hlfail none of three criteria met \\
Record-edit knockouts & \hlfail no combined signature met \\
\hlgrouprow \multicolumn{2}{@{}l}{\emph{Wider 7B searches, generation accuracy}} \\
Single sites & \hlfail 1 of 8 cells, 4 needed \\
Layer bands & \hlfail 1 of 5 cells, 3 needed \\
\bottomrule
\end{tabular}
}
\caption{\textbf{Interventions constrain explanations.} Counts by test, with each group's measure. Shading marks restoration and unmet criteria. $\ast$:~exploratory, $\dagger$:~post hoc.}
\label{tab:interventions-body}
\vspace{-5pt}
\end{table}

Following causal mediation and activation-patching practice~\citep{vig2020causal,meng2022locating,geva2023dissecting,zhang2024best}, we test whether transplanting a working internal state restores composition and whether a shared site, a few heads or successor access explain the failure, against matched controls on separate selection and report inputs (Table~\ref{tab:interventions-body}, Appendix~\ref{sec:app-where}).

Exploratory per-cell transplants raise generation accuracy by at least 10 points in 19 of 30 smaller-base cells, 18 of them also over a wrong-skill source. A post hoc logit-difference proxy at one fixed site per base, scored without rerunning generation, measures a different outcome rather than filtering the 19 cells, and no specified criterion identifies a shared local mechanism.

Mean ablation of selected readouts changes no cell, early-layer key knockouts restore none and small head sets meet no base-level criterion. Blocking all reads of successor keys keeps $A$ lost but reproduces the position-gap layout and cannot separate successor content from position. Restoration shows an intervention effect, not that the natural failure follows that pathway, so the tested interventions did not identify a local mechanism sufficient to explain the loss. Distributed interactions, untested sites and unfaithful interventions remain possible, and no behavioural finding above rests on localisation.

\noindent\textbf{Finding 2.} Proficiency does not certify insertion robustness, and insertion robustness does not certify compatibility. For layout-augmented skills, trained successors still interfere on two of four digit bases where a block of matched length and per-token norm does not. On two Qwen bases, success with the partner rows used in training largely fails to extend to other proficient rows of the same operations (post hoc). Interventions constrain explanations, but the tested interventions did not identify a local mechanism sufficient to explain the failure.

\section{Repair Can Compromise Skill Reuse}
\label{sec:repair}
\textbf{RQ3} asks whether repairing a composition preserves reusable skills: the edited skill alone and combinations absent from repair training.

\subsection{Repair Accuracy and Skill Retention}
\label{sec:repair-ablation}

We repair a failed composition by updating only the 20 embedding rows of its predecessor $A$ or successor $B$ using $N$ composed examples. A \emph{plain} edit optimises those examples alone. An \emph{anchored} edit also rehearses the edited skill on the same inputs under its original task, using $2N$ examples. Rehearsal is a standard retention strategy~\citep{robins1995catastrophic}, and here it tests whether composed accuracy characterises the edited module.

At eight composed labels, the edited skill's median standalone accuracy over pairs is 88.9\% to 99.0\% after anchored edits and 0.9\% to 23.0\% after plain edits across bases, skill seeds and edit sides, excluding 7B seed 1 (Figure~\ref{fig:fig_retention}). The contrast holds for both edit sides, including 7B seeds 2~and~3. On 20 interfering Qwen2.5-0.5B-Instruct record-editing pairs, a plain $B$ edit lowers $B$ alone by a median 86.0 to 100 points across label budgets and an anchored edit by at most 0.3~points.

These retention results use additional supervision, and no retention measurement matches an equal-budget comparison (Appendix~\ref{sec:app-repair}). The retention benefit of rehearsal at equal total supervision is therefore not quantified, which limits the method comparison but not the observation that composed accuracy does not certify retention.

More supervision can repair a pair while retaining its skills. On skill seed 1 of the three smaller digit bases, anchored successor edits reach composed accuracy of at least 90\% by 64 labels on 27, 20 and 25 of 27 pairs on 0.5B, 1.5B and Gemma-3-1B. At 1024 labels, 26 of 27 pairs on each base reach 95\%, and at the largest budgets no constituent of any pair drops more than 5 points.

Original-task retention is not retention in every earlier context. Anchored $B$ edits show no significant composition advantage over anchored $A$ edits (no equivalence test), and at eight labels on skill seed 1, including the failed 7B skill, an anchored $A$ edit lowers $A$ followed by \sk{ID} by a median 9.1 to 31.7 points across bases. Successor edits leave prompts omitting that successor unchanged by construction, a structural control rather than transfer (Appendix~\ref{sec:app-transfer}, Remark~\ref{rem:causal}).

\subsection{Repair Accuracy and Transfer}
\label{sec:repair-tuned}
\begin{table}[t]
\centering
\footnotesize
\setlength{\tabcolsep}{2pt}
\hlclip{\begin{tabular}{@{}lcc@{}}
\toprule
\hlgrouprow \multicolumn{3}{@{}l}{\emph{(a) Eight composed labels per pair}} \\
\hlheadrow \hlhd{Repair} & \hlhd{Composed} & \hlhd{Edited skill} \\
\midrule
\hlctrlrow No repair & 0.0$^\dagger$ & -- \\
Plain edit & 4.0--52.3 & \hlfail 1.2--23.0 \\
Anchored edit & 6.3--70.0 & \hlpass 88.9--98.9 \\
SPoT pair prompt & 9.3--69.0 & kept \\
Per-pair connector & 0.7--15.7 & kept \\
\midrule
\hlgrouprow \multicolumn{3}{@{}l}{\emph{(b) Transfer at 256 labels per training pair}} \\
\hlheadrow \hlhd{Repair} & \hlhd{Trained pairs} & \hlhd{Held out} \\
\midrule
\hlctrlrow No repair & -- & 5.7 \\
Anchored edit & 90.4 & \hlfail 10.4 \\
Connector per $B$ & 60.3 (82.9) & \hlfail 9.3 (10.6) \\
Connector, all pairs & 52.3 (72.4) & \hlfail 7.7 (9.0) \\
\bottomrule
\end{tabular}
}
\caption{\textbf{Composed accuracy, retention and transfer diverge across repairs.} (a)~Medians (\%) at eight labels, edit sides pooled, ranges over seed 1 of the three smaller bases and 7B seeds 2 and 3 ($\dagger$:~0.5B only). (b)~Means (\%) over 27 folds, or three nine-pair folds for the all-pairs connector, on one Qwen2.5-0.5B skill library, twice the epochs in parentheses.}
\label{tab:repair-summary}
\vspace{-5pt}
\end{table}

We compare edits with adapted SPoT, which tunes a 40-row pair prompt initialised from both skills~\citep{vu2022spot}, and with 20-row connectors that leave the library unchanged, trained per pair or shared across training pairs. Both use $N$ composed examples. Learning rates are selected on evaluated seed-1 pairs and then fixed, which is not held-out model selection, and shared connectors reuse the edit rate (Appendices~\ref{sec:app-tuned} and~\ref{sec:app-transfer}).

Anchored editing shows no consistent advantage over tuned SPoT (Table~\ref{tab:repair-summary}a). At eight labels on the smaller bases' first seeds, two-sided $p$ values are 0.11 to 0.89, and on 7B seeds 2 and 3 SPoT has the higher median without a significant difference. Anchored editing exceeds the per-pair connector in ten of 12 base and seed rows, but with twice as many examples. Matched at eight examples and 50 steps, tuned SPoT exceeds anchored editing on 16 to 24 of 27 pairs in each row (post hoc).

Repairs transfer little to unseen partners in the one tested Qwen2.5-0.5B skill library. In 27 leave-one-predecessor-out folds at 256 labels per training pair, anchored successor edits reach 90.4\% on trained pairs but 10.4\% on the held-out predecessor, against 5.7\% unrepaired, and only 4 of 27 folds gain at least 10 points (Table~\ref{tab:repair-summary}b, Figure~\ref{fig:partner-transfer}). Shared connectors transfer no better. With twice the epochs, held-out accuracy is 10.6\% for a connector shared by a successor's training predecessors and 9.0\% for one shared by all training pairs (the latter post hoc). Because shared connectors fit their training pairs only partly and the tests cover one 0.5B skill library, whose three repair-training seeds resample rather than replicate it, these results do not support a generally compatible successor or shared connector.

\noindent\textbf{Finding 3.} Repaired-pair accuracy certifies neither retention nor transfer. Plain edits can erase the edited skill, rehearsal keeps it without a consistent composition gain, and repairs on one Qwen2.5-0.5B library transfer little to held-out partners.

\section{Implications for Agent Skill Libraries}
\label{sec:agent-implications}

Agent libraries mostly use text or code skills (Section~\ref{sec:related}), so our parametric results motivate tests for them rather than establish their outcomes.

\paragraph{Admission and execution.}
A standalone length-8 gate would admit 70 of our 72 digit skill seeds (Table~\ref{tab:app-acceptance}), yet every directed pair still interferes in one call even with identity and order given. Admission tests could therefore cover ordered pairs on the intended inputs and layouts, since layout alone can suppress a skill (Section~\ref{sec:diag-layout}), and send untested or failing pairs to separate calls.

\paragraph{Replacement and updates.}
Plain repairs raise composed accuracy but largely erase the edited skill, and on one 0.5B library repairs transfer little to unseen partners (Section~\ref{sec:repair}). Versioning with non-regression replay~\citep{ma2026skillv} could address this if it covers the original task, earlier combinations and unseen partners. The replacement test (Section~\ref{sec:diag-partner}) suggests retesting skills trained next to a replaced skill. Appendices~\ref{sec:app-cost} and~\ref{sec:app-agent-implications} cover pair-specific serving cost and agent-level tests.

\section{Conclusion}
\label{sec:conclusion}

Independent skill acquisition does not establish that a library can reliably combine, replace and update its skills. In our controlled soft-prompt setting, skills that work alone and tolerate neutral insertions can still interfere. A post hoc test on two digit bases further shows that compatibility learned with one skill implementation need not transfer to another proficient implementation of the same operation. Repair can improve a composition while impairing the original skill, and retaining standalone performance does not establish transfer to unseen combinations. Rehearsal and additional supervision can support accurate compositions that preserve both skills. Internal interventions leave the natural failure mechanism unresolved. These model-level findings motivate evaluating skill libraries through supported combinations, replacement implementations and retained behaviours. For future self-evolving agents, the central question is whether acquiring and revising skills expands reliably reusable capabilities while preserving the behaviours on which existing compositions depend.
\unskip\label{body-end}

\section*{Limitations}
\label{sec:limitations}
\paragraph{Scope.}
Our evidence comes from soft-prompt skills on synthetic digit tasks and controlled record edits on a few open models, with skill identities and order supplied. It characterises compatibility inside the model and does not test agent-level selection and planning, longer skill chains, chain-of-thought execution or other skill representations.

\paragraph{Evidence.}
The controls and interventions narrow the causes of incompatibility without identifying the failure mechanism, and analyses marked post hoc or exploratory are descriptive (Appendix~\ref{sec:app-setup}).

\section*{Ethical Statement}
\label{sec:ethics}
The digit tasks are generated programmatically, and the record-editing tasks use synthetic records built from fixed lists of common names, cities and integer totals, so the study uses no personal data and involves no human annotators or participants. We use the Qwen2.5, Gemma-3, Llama-3.2 and Mistral models and the Skill Neologisms code under their licences and terms of use, for research only, on four NVIDIA H100 80GB GPUs (Appendix~\ref{sec:app-setup}). The repository linked in the abstract holds our code, including the code that generates every evaluation item, and the trained skills and digit checkpoints will be released with the paper where the applicable terms permit distribution. The 0.5B digit adapter released with Skill Neologisms states no licence (Appendix~\ref{sec:app-setup}).

This work studies how independently learned soft-prompt skills combine and how repairs change them. Its results concern the tested interface and do not show that any deployed skill library is safe or reliable. Editing a skill can silently change its behaviour in contexts that were not retested, so a repaired skill should pass regression tests on its original task, earlier combinations and new partners before it replaces the original. Whether such checks improve a deployed agent remains to be evaluated.

\bibliographystyle{acl_natbib}
{\exhyphenpenalty=1000 \bibliography{refs}\par}

\clearpage
\raggedbottom
\appendix
\renewcommand{\topfraction}{0.9}
\renewcommand{\bottomfraction}{0.8}
\renewcommand{\textfraction}{0.07}
\renewcommand{\floatpagefraction}{0.8}
\renewcommand{\dbltopfraction}{0.95}
\renewcommand{\dblfloatpagefraction}{0.8}
\setcounter{topnumber}{4}
\setcounter{dbltopnumber}{3}
\setcounter{totalnumber}{6}

\newcommand{\appFloatOrPending}[5]{%
  \IfFileExists{tables/#1_float.tex}{\input{tables/#1_float.tex}}{%
  \IfFileExists{tables/#2_float.tex}{\input{tables/#2_float.tex}}{%
  \IfFileExists{figures/#1_float.tex}{\input{figures/#1_float.tex}}{%
  \IfFileExists{figures/#2_float.tex}{\input{figures/#2_float.tex}}{%
  \PackageError{SkillCompatibility}{Missing supplementary float: #1}{Restore the corresponding float source before compiling.}}}}}}

\newcommand{\appTocLine}[1]{\noindent\makebox[1.4em][l]{\ref{#1}}\nameref{#1}\dotfill\pageref{#1}\par}
\twocolumn[{%
\begin{minipage}{\textwidth}
\section*{Appendix Contents}
\noindent\textit{Measurement definitions, experimental details, and supporting evidence}\par\smallskip
{\small
\noindent\begin{minipage}[t]{0.47\textwidth}
\appTocLine{sec:app-setup}
\appTocLine{sec:app-revisit}
\appTocLine{sec:app-diag}
\appTocLine{sec:app-c1}
\end{minipage}\hfill
\begin{minipage}[t]{0.47\textwidth}
\appTocLine{sec:app-where}
\appTocLine{sec:app-repair}
\appTocLine{sec:app-agent-implications}
\end{minipage}\par}
\section*{Guide to the Supplementary Evidence}

\noindent
The supplement follows the order of the main text: Appendix~\ref{sec:app-setup} supports Section~\ref{sec:setup}, \ref{sec:app-revisit} supports RQ1 (Section~\ref{sec:revisit}), \ref{sec:app-diag} to~\ref{sec:app-where} support RQ2 (Section~\ref{sec:diag}), \ref{sec:app-repair} supports RQ3 (Section~\ref{sec:repair}) and \ref{sec:app-agent-implications} extends Section~\ref{sec:agent-implications}. Table~\ref{tab:app-evidence-map} links each claim to its supporting comparison, detailed results and inference boundary.

\medskip
\begingroup
\small
\setlength{\tabcolsep}{4pt}
\renewcommand{\arraystretch}{1.0}
\noindent
\begin{tabular}{@{}>{\raggedright\arraybackslash}p{0.18\textwidth}>{\raggedright\arraybackslash}p{0.27\textwidth}>{\raggedright\arraybackslash}p{0.34\textwidth}>{\raggedright\arraybackslash}p{0.15\textwidth}@{}}
\toprule
\textbf{Main-text claim} & \textbf{Decisive comparison} & \textbf{Scope of inference} & \textbf{Supporting evidence} \\
\midrule
\textbf{Skills and evaluation are defined before comparisons.}\newline
Section~\ref{sec:setup}
& Define skills, ordered targets, acceptance, the output-level presence measures and the statistical units.
& Presence measures record output effects, not internal states. Pairs share constituent skills.
& Appendix~\ref{sec:app-setup}. \\
\addlinespace[0.4em]
\textbf{RQ1: Proficiency is not compatibility.}\newline
Section~\ref{sec:revisit}
& Compare one-call and two-call execution on matched inputs. Inspect constituent acceptance, directed pairs and the operations visible in failed outputs.
& The gap concerns the tested skills and execution settings. Output omissions do not identify a common failure mechanism across digit and record tasks.
& Appendices~\ref{sec:app-acceptance-scope}, \ref{sec:app-revisit}. Tables~\ref{tab:phenomenon}, \ref{tab:app-accept-sens}. Figure~\ref{fig:app-matrix}. Tables~\ref{tab:app-failure}, \ref{tab:app-nl-mech}. \\
\addlinespace[0.4em]
\textbf{RQ2: Layout alone can suppress a skill.}\newline
Section~\ref{sec:diag-layout}
& Vary inserted content, position-ID gaps and skill-block length while keeping the earlier operation's target fixed.
& $A$-present records an output effect. Position gaps are artificial controls. The block-length comparison covers one base and one seed.
& Appendices~\ref{sec:app-d1}, \ref{sec:app-ll}. Figure~\ref{fig:successor}. Table~\ref{tab:app-ll}. \\
\addlinespace[0.4em]
\textbf{RQ2: Robust skills still interfere.}\newline
Section~\ref{sec:diag-compat}
& Compare augmented skills under a trained successor and random blocks matched in row count and per-row norm, with repeated random directions.
& On augmented Qwen 0.5B and 1.5B, length and row norms alone are insufficient. The count rule was amended. No particular learned direction or pathway is identified.
& Appendices~\ref{sec:app-l1}, \ref{sec:app-compat}. Table~\ref{tab:app-l1}. Figure~\ref{fig:compat-detail}. \\
\addlinespace[0.4em]
\textbf{RQ2: Learned blocks need not be interchangeable.}\newline
Section~\ref{sec:diag-partner}
& Train next to frozen learned partners, then replace the partner rows with independently learned rows of the same operations.
& Two Qwen bases, one recipe and a post hoc replacement test. The drop is consistent with compatibility forming between the specific blocks met in training.
& Appendices~\ref{sec:app-neo-reverse}, \ref{sec:app-l2}. Tables~\ref{tab:l2-body}, \ref{tab:app-l2}. \\
\addlinespace[0.4em]
\textbf{RQ2: Documents help some skill combinations.}\newline
Section~\ref{sec:diag-carrier}
& Compare documents with soft blocks, then vary worked examples, names, length and added trained or norm-matched random blocks.
& The pair-class split is post hoc. Constituent reliability and information differ. At 7B, the 29-pair overall evaluation and 30-pair class analysis have distinct denominators.
& Appendices~\ref{sec:app-c1}, \ref{sec:app-document-examples}, \ref{sec:app-c2}. Tables~\ref{tab:carrier}, \ref{tab:app-c1}. Figure~\ref{fig:app-carrier}. \\
\addlinespace[0.4em]
\textbf{RQ2: Interventions constrain explanations.}\newline
Section~\ref{sec:diag-probes}
& Compare activation replacements with wrong-skill sources and common sites. Test restricted head sets and selective access to the other skill.
& Restoration is an intervention effect. Failed local tests do not imply a distributed mechanism. Full knockouts can match a position-gap control.
& Appendix~\ref{sec:app-where}. Tables~\ref{tab:transplant}, \ref{tab:app-nb}. Figure~\ref{fig:fig_ha}. \\
\addlinespace[0.4em]
\textbf{RQ3: Repair can compromise skill reuse.}\newline
Section~\ref{sec:repair}
& Compare plain and anchored edits on composition and standalone retention. Separately evaluate tuned baselines, unseen partners and other contexts.
& Anchoring adds examples. Standalone retention does not ensure that other compositions are intact. Held-out-predecessor transfer covers only Qwen 0.5B. Sharing one connector across 18 training pairs is a post hoc check.
& Appendix~\ref{sec:app-repair}.\newline Appendices~\ref{sec:app-tuned}, \ref{sec:app-transfer}.\newline Tables~\ref{tab:app-ablation-seeds}, \ref{tab:app-transfer}. \\
\addlinespace[0.4em]
\textbf{Implications for agent skill libraries are prospective.}\newline
Section~\ref{sec:agent-implications}
& Map each measured property to an admission, execution or update decision for agents that adopt this interface.
& Model-level evidence only. No agent was built or evaluated.
& Appendix~\ref{sec:app-agent-implications}. Figure~\ref{fig:agent-library-scope}. \\
\bottomrule
\end{tabular}
\endgroup

\begingroup
\captionsetup{hypcap=false}
\captionof{table}{\textbf{Main claims and their supporting evidence.} The scope column states what each comparison can establish.}
\label{tab:app-evidence-map}
\endgroup
\medskip

\end{minipage}%
}]

\section{Experimental Design and Reproducibility}
\label{sec:app-setup}

Skill identities and execution order are supplied by the evaluator, two-call conditions follow a fixed sequence, and the same conventions apply to subsequent controls unless a local protocol states a change. Throughout, $A$ is the first operation and $B$ its successor. A \emph{unit} is a trained skill and its seed. A \emph{cell} is a directed pair and seed on one base model. Units and cells identify the level of aggregation, but share skills and evaluation items. They are not independent experimental replications.

\subsection{Digit Tasks and Independent Skill Learning}

The six digit skills preserve string length. \sk{SHIFT\_RIGHT} (SR) moves the last digit to the front. \sk{SHIFT\_LEFT} (SL) moves the first digit to the end, and \sk{SWAP\_PAIRS} (SWAP) exchanges adjacent digits. \sk{INVERT\_POLARITY} (IP) maps even digits to 1 and odd digits to 0, \sk{MOD3} applies the remainder modulo 3 to each digit, and \sk{MIRROR} maps $d$ to $9-d$. The core pair is \sk{SR}\sk{IP}. The digit base is further pretrained on \sk{ASC}, \sk{DESC}, \sk{ADD} (increment modulo 10), \sk{SUB}, \sk{POLARITY} (the complement of \sk{IP}), \sk{REVERSE} and \sk{ID}.

Following the Skill Neologisms recipe~\citep{berthon2026skill}, a skill consists of 20 new embedding rows optimised by cross-entropy while all other parameters remain frozen. Each training example combines that skill with at most two pretrained operations, excluding \sk{ADD}. Thus the skill encounters neighbours such as \sk{ID}, \sk{REVERSE} and \sk{POLARITY}, but never another newly learned skill. Training uses 99,999 examples, three epochs and a learning rate of 0.005. We train three seeds per skill and hold out lengths 5, 7 and 9. In every seed, all 20 rows start from the same fixed vector, which also defines the \sk{INIT} control. For the Qwen and Llama bases, this vector is the mean embedding of the seven pretrained operation names, each first averaged over its sub-tokens. For the Gemma and Mistral bases, whose tokenisers map these names to the unknown token, it is the unknown-token embedding. The skill seed changes only which examples are sampled. Example order is shuffled with the trainer's fixed default seed of 42 in every run, for digit and language skills alike. Gemma prompts include the BOS token used during skill training. Qwen, Gemma and Mistral tokenise individual digits separately.

The six skills define 30 directed pairs with distinct constituents. We exclude \sk{SR}\sk{SL} and \sk{SL}\sk{SR}, whose composition is the identity, and \sk{IP}\sk{MOD3}, for which the second operation has no observable effect. The remaining 27 pairs include 18 commuting pairs. Excluding these degeneracies makes it possible to distinguish successful composition from outputs that omit an operation.

\subsection{Language Tasks and Alternative Representations}

Language inputs have the form \texttt{name: G S | city: C | total: T}, where G and S are given and family names, C is a city and T is an integer. \sk{NL\_NAME\_SWAP} writes the surname first, followed by a comma and the given name. \sk{NL\_NAME\_INITIAL} abbreviates the given name, and \sk{NL\_SURNAME\_CAPS} capitalises the surname. \sk{NL\_ADD\_SHIPPING}, \sk{NL\_DOUBLE\_TOTAL} and \sk{NL\_ROUND\_DOWN} add 5, double the total and round it down to a multiple of 10. The 30 directed pairs comprise 18 different-field pairs, 6 same-name-field pairs and 6 same-total pairs. Only the total pairs are non-commuting.

Soft language skills contain 20 new embedding rows and are trained with three skill seeds. Fewer than two candidate native steps editing the city or total passed the preliminary gate. We therefore followed the specified fallback and trained each skill alongside the \sk{ID} filler only. Unlike the digit recipe, this training supplies no functional pretrained neighbours. Language consequently tests the reach of the composition question, while differences from digits also include the training context.

The soft representation prefixes the record with \texttt{Steps: [A][B]}. The document representation supplies one \texttt{SKILL.md} document per invoked skill in the user turn, followed by \texttt{Steps: nameA, then nameB}. A document contains a name, a one-sentence description, a procedure and two worked examples, varied across three document draws. At 7B, a pair requires approximately 405 prompt tokens as documents and 101 as soft blocks. The library condition loads all six documents. The plain-text condition uses one sentence per skill. All representations are evaluated on the same 300 records derived from the digit task's length-8 strings.

A pair is eligible only when both constituents pass the single-skill gate in that seed. \sk{NL\_SURNAME\_CAPS} fails as a soft skill in every seed on Qwen2.5-0.5B-Instruct and 1.5B-Instruct. Consequently, 20 of 30 pairs are eligible on those bases and all 30 on 7B. Pairs containing an ineligible constituent are excluded from the specified language claims.

\subsection{Observable Effects and Paired Evaluation}

\label{sec:app-presence}

Exact composition and preservation of an individual skill answer different questions. For $A$ followed by a block, \emph{$A$ present} means that the output is $A(x)$ or, when the block has a known operation, that operation applied to $A(x)$. With a trained successor, both $A(x)$ and $B(A(x))$ count. This is an observable effect measure: it does not establish that an intermediate representation explicitly encodes $A(x)$, and it can exceed exact composition accuracy when only $B$ is omitted.

Rule-selected example outputs in the released code (masked decoding, first item of skill seed 1) show $A$ present in none of 12 digit cases beyond the core pair, in line with the rate after a trained successor, below 20\% in most pair and seed cells (Appendix~\ref{sec:app-d1}). Language examples are scored on complete records, which is stricter than the field flags of Section~\ref{sec:setup}, since an output can show $A$'s edit yet count as $A$ absent when $B$'s edit is wrong.

Each digit condition evaluates 300 length-8 strings, paired by item index across conditions within a seed. Decoding is greedy and excludes newly added skill-token IDs from the output vocabulary, as detailed in Appendix~\ref{sec:app-mask}. Because the digit base can continue generating, exact-match scoring uses the first $|y|$ characters, where $y$ is the gold output. Analyses of overlong or malformed generations also inspect the generated output.

The reported digit two-call control uses the diagnostic subset on which the gold composition, $B(x)$, $A(x)$ and $x$ are pairwise distinct, so an omitted operation cannot receive credit by coincidence. Table~\ref{tab:phenomenon} reports one-call means on the same subset, 6.4\%, 1.3\%, 0.9\% and 2.7\% for Qwen2.5-0.5B, 1.5B, 7B and Gemma-3-1B. One-call means elsewhere, as in Figure~\ref{fig:app-matrix} and Tables~\ref{tab:app-accept-sens} and~\ref{tab:app-failure}, use all evaluation items (6.8\%, 1.8\%, 1.1\% and 3.1\%), so the two denominators should be distinguished.

The independence reference is the product of the constituent single-skill accuracies. The interference test averages a one-sided exact binomial test over Jeffreys-posterior draws of these accuracies. A pair is counted as interfering when the effect holds in at least two of its three skill seeds, using each seed's own constituent accuracies. Consequently, an unreliable constituent lowers the reference for its pairs. This procedure does not model the dependence between single and joint runs on shared items. Its significance levels are approximate.

\subsection{Prerequisite Competence and Acceptance}

\label{sec:app-acceptance-scope}

Two checks separate a failure to learn the task from a failure to combine learned skills: competence of the digit base on its pretrained operations, and competence of each new skill on held-out inputs. We use the released Qwen2.5-0.5B digit base of Skill Neologisms and further pretrain Qwen2.5-1.5B, Qwen2.5-7B and Gemma-3-1B with the same operation set and configuration~\citep{qwen2024qwen25,gemma2025gemma3}. A rerun on Qwen2.5-0.5B differs from the released base by at most 4.0 points on each evaluation set.

The Qwen2.5-1.5B and 7B float16 runs retained their best, early checkpoints after evaluation loss became NaN. Gemma-3-1B overflowed in float16 and was therefore trained in bfloat16, completing its schedule without a NaN loss. The retained Qwen bases nevertheless achieve 99.0\% and 100\% accuracy on single pretrained operations at length 8. Their two-operation accuracies at lengths 6 and 8 are 96.0\% and 88.0\% on 1.5B, and 94.0\% and 81.0\% on 7B, against 96.0\% and 90.0\% on the released 0.5B base. These sets draw each operation independently, so 11 of the 100 items at length 6 and 19 of the 100 at length 8 repeat one operation, a combination that base pretraining, drawing without replacement, never produces. In a post hoc descriptive rescore of the 89 and 81 items with two different operations, accuracy at lengths 6 and 8 is 100\% and 100\% on 0.5B, 100\% and 97.5\% on 1.5B, and 97.8\% and 88.9\% on 7B.

A post hoc descriptive check from the same skill runs asks whether a newly learned skill combines with its six training neighbours (every pretrained operation except \sk{ADD}). Masked one-call accuracy at length 8 (100 items per cell) has median 100\% in both orders on each of the four bases. The order \texttt{[S][P]x} reaches at least 90\% in 102, 102, 93 and 108 of 108 skill, seed and operation cells on 0.5B, 1.5B, 7B and Gemma-3-1B, and \texttt{[P][S]x} in 99, 82, 92 and 107, so 26 of 108 cells on 1.5B fall below 90\% when the pretrained operation comes first. On 7B seeds 2 and 3 the counts are 63 and 62 of 72. The 15 7B cells below 90\% in the first order come from \sk{MOD3} in seed 1, \sk{MIRROR} in seed 2 and \sk{INVERT\_POLARITY} in seed 3. Base pretraining includes compositions of one to three operations, and each skill was trained with these neighbours in both orders. The base check therefore establishes competence in composing operations learned together, and the post hoc neighbour check is consistent with it. Neither concerns operations learned separately, and neither separates familiarity with the pair from whether the partner is pretrained. The same learned-skill failure on the released and fully trained Gemma bases also argues against incomplete pretraining as a sufficient explanation, although the unequal convergence histories remain a limitation.

The original acceptance rule requires mean accuracy of at least 90\% over held-out lengths 5, 7 and 9. Because the Qwen digit bases themselves reach only 81.0\% to 84.8\% on single pretrained operations at length 9 (91.4\% on 7B), we additionally report a revised rule using lengths 5 and 7. This change was made after the runs and is a protocol deviation.

Table~\ref{tab:app-acceptance} reports both rules, using unmasked evaluation with each skill installed alone. Each skill, base and seed uses the checkpoint of the corresponding six-skill library run, installed alone by \texttt{code/digits/eval\_skills.py} rather than with the library by \texttt{code/digits/eval\_matrix.py}. A post hoc check finds skill tokens in the scored outputs only on the excluded Llama-3.2-1B, whose counts are the same under masking, and in 5 of the 200 length-7 outputs of SR on Qwen2.5-1.5B seed 2. Masking would raise that skill's held-out mean from 88.7\% to 90.0\% and pass it under the revised rule. On 7B, \sk{MOD3} in seed 1 achieves only 3.0\% at length 8, compromising 9 of that seed's 27 pairs, so later repair comparisons also report seeds 2 and 3. The four other 7B skill seeds that fail the revised held-out rule still reach at least 89.0\% at length 8. A length-8 gate alone would thus admit 70 of the 72 skill seeds on the four included bases, rejecting only that \sk{MOD3} seed and \sk{INVERT\_POLARITY} in 7B seed 3 (89.0\%). The primary interference matrices and repair tables do not filter out failed skill seeds.

\input{tables/app-acceptance_float.tex}

To assess whether weak constituents explain the interference result, Table~\ref{tab:app-accept-sens} restricts the analysis to cells whose two skills pass each acceptance rule, reporting masked and unmasked decoding separately. Besides the two rules, a length-8 rule requires both skills to reach 90\% at the training length. Masked one-call accuracy remains significantly below the independence reference after Holm correction within the retained set in every retained cell except at most one per base, under every rule. On 7B, excluding seed 1 with its failed \sk{MOD3} raises two-call accuracy from 88.2\% to 98.1\% and the independence reference from 88.5\% to 98.7\%. A stricter criterion taken from the language setting, one-call accuracy at least 10 points below two calls and significantly lower in most seeds, keeps all 27 pairs on 1.5B, Gemma-3-1B and 7B and 26 of 27 on 0.5B. The exception is \sk{MIRROR} followed by \sk{INVERT\_POLARITY}, which reaches 97.0\% and 100\% against 100\% in seeds 2 and 3. The original rule retains no 0.5B pair cell, and an empty accepted subset constitutes evidence neither for nor against interference. The phenomenon therefore persists among the reliable constituents, although the accepted subsets differ across rules.

\input{tables/app-accept-sens_float.tex}

\subsection{Why Novel Output IDs Are Masked}

\label{sec:app-mask}

Qwen2.5-0.5B, Qwen2.5-1.5B and Gemma-3-1B tie their input and output embeddings, so a newly learned skill row can become an output token even though the required answer contains only digits. Masking therefore forbids every skill-token ID during decoding, 120 for the six-skill library (20 per skill). The lower block of Table~\ref{tab:app-accept-sens} compares masked and unmasked decoding over the 81 digit pair and seed cells per base.

Without masking, the tied bases emit skill tokens in up to 95.7\% of composed outputs and the untied Qwen2.5-7B in none. Unmasked single-skill accuracy collapses on the two smaller Qwen bases and lowers their independence references, so only 0 of 27 and 13 of 27 pairs fall below them, compared with 27 of 27 under masking. Gemma-3-1B, which emits skill tokens less often and keeps an unmasked single-skill accuracy of 93.9\%, and 7B have 27 of 27 interfering pairs under either condition. The mask excludes only the new IDs. Every unmasked output that emits no skill token agrees with its masked counterpart through the first end token, and all 7B outputs are identical under the two conditions. With one skill installed alone, as in the acceptance runs, only its 20 IDs are added, no length-8 output of that skill contains a skill token, and unmasked accuracy equals masked accuracy. The unmasked comparison on the two smaller Qwen bases is therefore dominated by skill-token emission and cannot cleanly measure composition.

\subsection{Limits of Recipe Transfer across Model Families}
\label{sec:app-ms}

The attempted Llama and Mistral extensions ask whether the training recipe produces reliable constituents at all. Their completed outputs are retained, but low joint accuracy on these models cannot establish the composition result.

On Llama-3.2-1B~\citep{grattafiori2024llama3}, whose tokenizer groups one to three digits, only 1 of 18 skill seeds passes acceptance (Table~\ref{tab:app-acceptance}), and training loss remains near 0.2, compared with 0.012 on Qwen2.5-0.5B. Mean joint accuracy is 0.6\%, but no pair falls below its already low independence rate. We therefore exclude this base from composition claims. This outcome identifies a failure of the recipe on that model. It does not isolate tokenisation as the cause.

The Mistral-7B-v0.3~\citep{jiang2023mistral} extension was specified before its code changes and runs, with the same digit operations, thresholds and rules, a bfloat16-pretrained digit base, six skills with three seeds, their augmented versions and the layout controls. Its base passes the base-level gate, but none of the six seed-1 skills passes revised acceptance, which triggers the skill-level stop rule (lower block of Table~\ref{tab:app-acceptance}). Eight of the twelve units reaching 90\% at length 8 fail on held-out lengths 5 and 7, so the failure lies in learning sufficiently generalisable skills under the specified recipe, not in the base's prerequisite competence. Composition jobs had already been queued and completed before the acceptance check was applied. Their outputs are retained descriptively, but the specified stop rule excludes them from composition claims. Mistral contributes a documented recipe-transfer limitation, not a successful third-family replication.

\subsection{Numerical Precision and Sampling Conventions}

Digit evaluation uses one fixed item draw, each base one random-row draw shared across skill seeds, and augmentation distractors a separate fixed draw. \sk{INIT} rows involve no random draw, since they copy the fixed vector from which every trained skill block starts. Localisation configurations are selected on 300 items and evaluated on a disjoint pool of 500. Insertion lengths retain their specified units: on Qwen, one \sk{ID} occupies two positions and \sk{REVERSE} four, whereas the neutral filler is the token `the', so a number of filler copies need not equal the number of positions. Head interventions use eager attention, in bfloat16 on Qwen2.5-0.5B and float32 on 1.5B and 7B, because bfloat16 eager attention yielded single-skill accuracies of 52.2\% and 35.6\% on the latter bases, compared with 100\% under sdpa. The remaining probes and controls use sdpa. Reported seed ranges are minima and maxima, not confidence intervals. Pooled counts such as 22 of 22 or 74 of 81 are descriptive because cells share constituents, seeds and items.

\subsection{Analysis Status and Artefact Availability}

Here, \emph{pre-specified}, called \emph{logged} in the main text, means recorded in an internal time-stamped log before the relevant runs, rather than publicly registered before the study. Later experiments could be informed by earlier results, with the deviations listed below and post hoc checks identified explicitly. Numerical requirements are minima unless an upper bound is stated. As for any unmet threshold, failed local restoration does not identify a distributed natural mechanism.

The skill-acceptance rule was revised after the runs. The norm-matched digit rule was amended after insertion-control results were known, including the trained-successor values reused in that comparison. The document norm-matching thresholds, harm on 8 of 18 pairs or sparing on at most 2, were chosen from the trained-block counts when ordinary random blocks already harmed 1 of 18. The carrier comparison was split by pair class after the 7B results. Its specified all-pair outcome is reported alongside that split.

Transplant source-site and layer rules were set after initial runs, making that analysis exploratory. Treating cells with zero restoration from both attention and MLP outputs as unclassified was also a later interpretation. It changes the aggregate conclusion only on Qwen2.5-1.5B. When the Llama arm was excluded, the prediction study moved to held-out Gemma-3-1B and Qwen2.5-7B with a stricter success rule. Extensions to additional bases were fixed after earlier-base results but before their own runs, retaining the original numerical thresholds. Finally, the graded digit error attribution of Appendix~\ref{sec:app-failure} is a post hoc reanalysis. In the learned-partner study of Appendix~\ref{sec:app-l2}, the original-row evaluation is pre-specified, whereas the new-row check and the all-pairs connector check of Appendix~\ref{sec:app-transfer} are post hoc. Unsuccessful predictions remain part of the evidence reported here.

\paragraph{Licences and compute.} Qwen2.5-0.5B, 1.5B and 7B, their Instruct variants, Mistral-7B-v0.3 and the Skill Neologisms code use Apache 2.0, Gemma-3-1B uses the Gemma Terms of Use, and Llama-3.2-1B uses the Llama 3.2 Community License. The released 0.5B digit adapter states no licence. We use these artefacts for research. All experiments ran on four NVIDIA H100 80GB GPUs: further pretraining of the digit bases, training of every skill, composition repair and all reported evaluations, including the dedicated latency measurement in Appendix~\ref{sec:app-cost}, which used one of them.

\section{RQ1: Composition Across Skill Pairs}
\label{sec:app-revisit}

\begin{figure}[t]
\centering
\includegraphics[width=\columnwidth]{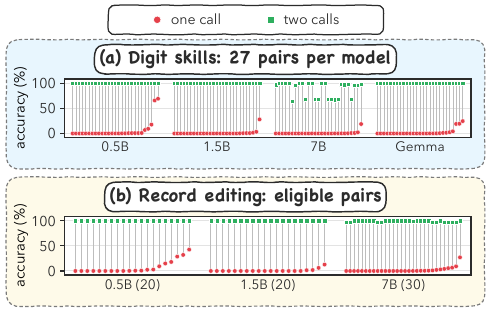}
\caption{\textbf{The composition gap extends across skill pairs.} One-call (red) and two-call (green) accuracy in percent per directed pair, averaged over three seeds.}
\label{fig:drop}
\end{figure}

This appendix supports Section~\ref{sec:revisit} (RQ1). Figure~\ref{fig:drop} plots each directed pair. Digit two-call controls use the diagnostic inputs of Table~\ref{tab:phenomenon} and record-editing points use eligible pairs. Some Qwen2.5-0.5B pairs compose in individual seeds, and the 7B averages include seed 1, whose \sk{MOD3} fails alone.

\subsection{Failure across Pairs and Execution Orders}
\label{sec:app-pair-order}

Failure extends beyond the core pair. Over the 27 non-degenerate directed pairs (Figure~\ref{fig:app-matrix}), mean masked one-call accuracy is 6.8\%, 1.8\%, 1.1\% and 3.1\% on Qwen2.5-0.5B, 1.5B, 7B and Gemma-3-1B. The best cell on each Qwen base is \sk{IP}\sk{SR}, the reverse of the core pair, reaching 69.3\% on 0.5B, so a few exceptions coexist with low aggregate accuracy.

\input{tables/app-matrix_float.tex}

Across the nine combinations of skill seeds, \sk{SR}\sk{IP} averages 0.3\% to 3.3\% per base and \sk{IP}\sk{SR} 11.4\% to 44.0\%. Because the reverse order varies across seed combinations, tests use the three matched seeds. They favour the reverse order after Holm correction in all three seeds on 0.5B and 1.5B, but in one of three on 7B and Gemma-3-1B. Across the 13 unordered non-degenerate pairs, significant order effects occur in at least two seeds for 4 pairs on 0.5B, 1 on 1.5B, none on 7B and 3 on Gemma-3-1B. Order sensitivity is therefore pair dependent, and is not itself new~\citep{yang2023tailor,radevski2026steering}.

For commuting pairs, retention is the fraction of identifiable outputs showing an operation divided by its single accuracy with all skills loaded. It is lower for the first skill on every base: 0.046 to 0.070, against 0.105 to 0.185 for the second. Appendix~\ref{sec:app-mask} gives the decoding sensitivity.

A pretrained separator helps little. Inserting \sk{ID} as \sk{X}\sk{ID}\sk{Y} significantly improves at least two seeds for no Qwen pair and 1 of 27 Gemma pairs, with mean changes of $-0.5$ to $-0.2$ points. In the core pair, \sk{ID} before, between or twice between the skills improves no seed on any base, so a familiar boundary marker alone is insufficient under this recipe.

\subsection{What Pooled Commuting Orders Conceal}
\label{sec:app-sn}

Pooling commuting orders can hide this asymmetry, because both token sequences have the same target. Skill Neologisms already studies sequential digit composition, including \sk{SHIFT\_RIGHT} and \sk{INVERT\_POLARITY} on Qwen2.5-0.5B, and reports the two orders together~\citep{berthon2026skill}. Our comparison concerns order resolution, not sequential composition itself.

We rerun that digit protocol on the released Qwen2.5-0.5B base. It shuffles the two orders, pools lengths 2 to 8 with fifty inputs per length, samples at temperature 1 and zeroes skill unembeddings, so its values are not comparable with our greedy tables. Inputs are shared across seeds, and item-level outputs were not stored. The published order-pooled curve falls from about 57\% at length 2 to about 21\% at length 8, a mean of about 33\% (read from Figure 5 of that paper). Our pooled scores of 49.4\%, 27.4\% and 48.0\% for skill seeds 1 to 3 lie in that range. Disaggregated over 184 $AB$ (shift first) and 166 $BA$ items per seed, $AB$ reaches 13.6\%, 13.6\% and 12.5\% and $BA$ 89.8\%, 44.0\% and 86.1\%. The $AB$ rates of seeds 1 and 2 coincide at every length, and seed 3 differs only at length 6. $BA$ is higher in all three seeds under Fisher's exact test ($p\le 2.7{\times}10^{-10}$), and the 95\% Wilson intervals of the two orders do not overlap, so pooling hides order-specific reliability. The distinction holds in three of three seeds with a larger item draw and, at the published sample size, in two of three seeds on Qwen2.5-1.5B, one of three on 7B and three of three on Gemma-3-1B.

Its language evaluation, two Skill-Mix skills in one GPT-5-graded text on Llama-3.2-3B-Instruct trained next to up to two other skills, differs from our exact-match record edits, whose specified fallback uses only \sk{ID} neighbours. We therefore reproduce only the digit protocol and treat our language results as neither a replication nor a contradiction of that study.

\subsection{What Digit Errors Reveal}
\label{sec:app-failure}

An output can fail exact composition while retaining one operation, but attributing every error to a dropped skill would go beyond the observations. We first assign masked outputs, pooled over three skill seeds, to the mutually exclusive classes of Table~\ref{tab:app-failure}, upper panel, taken left to right. Short outputs have fewer than eight digits, partial outputs match at least six target digits, and shifted outputs match a candidate shifted by one position. Chance compares the same generations with another item's candidates.

\input{tables/app-failure_float.tex}

For \sk{SR}\sk{IP}, 32.6\% to 54.3\% of outputs equal $B(x)$, omitting the first skill, and none equals $A(x)$. Across all pairs, these fractions are 8.5\% to 12.0\% and 0.8\% to 4.6\%, against chance at most 0.1\%. One-position shifts account for at most 0.2\% of outputs. Partial application exceeds chance only for commuting pairs, at 15.6\%, 8.3\%, 7.0\% and 11.6\% against 4.8\%, 4.6\%, 5.2\% and 3.5\% on 0.5B, 1.5B, 7B and Gemma-3-1B, and the released code reproduces the split by pair class. Most outputs match no class (51.3\% to 66.8\% per base), including the 27.2\% of 7B outputs that continue past eight digits, so the clean dropped-operation cases do not describe every failure.

A post hoc graded analysis (lower panel) assigns each output to its digit-wise closest candidate among $B(A(x))$, $B(x)$, $A(x)$ and $x$. Ties comprise 19.7\% to 33.8\% of outputs and are more frequent under chance. Conditional on a unique closest candidate, first-skill loss exceeds chance on every base and second-skill loss falls below it. With ties, second-skill loss is at or slightly above chance on three bases: 15.8\% versus 15.4\% on 0.5B, 17.2\% versus 13.7\% on 1.5B and 18.7\% versus 15.8\% on 7B. Pair counts use all outputs, and the first-loss excess leads in only 16 to 18 of 27 pairs (two-sided sign-test $p$ of 0.12 to 0.44 when masked and 0.0059 for unmasked 0.5B, approximating shared-skill pairs as independent). The evidence supports an aggregate tendency towards first-skill loss, not a significant pair-level law. Intervention evidence that $A$ is hard to preserve appears in Appendix~\ref{sec:app-d1}.

\subsection{Language Reverses the Omission Pattern}
\label{sec:app-nl-mech}

Language preserves the composition failure but reverses the dominant omission. The specified fingerprint, the share missing the earlier skill minus the share missing the later one, is $-30.2$, $-49.9$ and $-59.3$ points over the 60, 60 and 90 eligible pair-seed cells of Qwen2.5-Instruct 0.5B, 1.5B and 7B, and negative in 37, 46 and 68 cells. On 0.5B and 1.5B, one-call accuracy on eligible pairs, scored on the same records as two calls, is 7.6\% and 1.1\%, against 5.4\% and 0.7\% over all thirty pairs.

To decompose this, we recomputed lenient field-level application flags post hoc for all 300 records in each cell, reproducing the original scores on parsed outputs. An operation counts as applied when its own field shows the edit, even in a partially incorrect record. The decomposition is descriptive and does not identify a single-slot mechanism. As Table~\ref{tab:app-nl-mech} shows, the later skill is applied in 30.2\%, 15.9\% and 10.1\% of eligible outputs, against 60.3\%, 65.8\% and 71.5\% for the earlier skill, and the ordering holds within every pair class. Descriptively, each of the six skills is also applied more often first than second on every base when averaged over partners, ineligible ones included. The mean loss row gives standalone accuracy minus applied share, and the released code reproduces the per-skill rows.

\input{tables/app-nl-mech_float.tex}

Given the gold intermediate record in a separate call, five skills reach at least 99.9\% and \sk{NL\_SURNAME\_CAPS} 55.0\% to 97.2\%, so omission of a reliable successor is not simply an inability to process intermediate inputs. The earlier skill is not fully retained either. Outputs with neither edit are 26.7\%, 28.6\% and 21.4\%, mainly input copies on 0.5B and 1.5B (15.5\% and 18.6\% of all outputs). At 7B, 18.0\% of outputs lack the name or total field and fill entire pair-seed cells, and outputs with both fields show $A$ in 85.3\% and $B$ in 12.3\%.

\subsection{What the Language Controls Establish}
\label{sec:app-language-controls}

The controls test whether the reversed pattern reduces to prefix sensitivity, missing order cues or one representation. They exclude each only in part and select no single shared mechanism. Neither \sk{ID} nor ``then'' between soft blocks restores the later edit. Random rows hurt more before a language skill than after it, reversing the digit pattern. Yet on 7B, 20 random rows before a total skill cost it at most 2.9 points, while these skills are applied in only 0.2\% to 19.6\% of one-call outputs when second, so ordinary prefix sensitivity cannot account for those failures. Documents apply both edits on 91.7\% of 7B different-field outputs and apply $B$ more often than $A$ at 1.5B (73.7\% versus 52.8\%), but on 7B same-field pairs keep only the first edit in 46.2\% of outputs and only the second in 11.1\%. The asymmetry thus depends on representation and pair class, not on the soft representation alone.

Of the four specified language predictions, the first two require earlier-loss asymmetry in a third of a pair class and selective disruption by new blocks rather than by \sk{ID}, text or prose. Asymmetry holds in three, two and zero different-field pairs against thresholds of four, four and six (cluster bootstrap: two, two and zero), and in none of the six same-field commuting pairs on 7B, the only base where they are evaluable. The prefix test concerns the earlier skill alone over eighteen skill-seed cells, not the later edit in the composed prompt. In the prefix controls of Table~\ref{tab:app-nl-mech}, \sk{ID} before or after a block changes $A$'s presence by at most 0.6 points, random or untrained blocks largely preserve it on 0.5B and 7B, and a text instruction or neutral prose reduces it on 1.5B and 7B. Selective disruption therefore fails on 0.5B and 1.5B, and no 7B pair remains eligible, so neither prediction is supported. The textual partner instruction was unseen in the filler-only training recipe, and the planned native-step condition was not run after that fallback. The third prediction has no eligible commuting pair on any base, so its order test is uninformative about whether order effects exist. The fourth, on $A$'s own field in different-field pairs, holds in 7, 6 and 7 pairs against thresholds of 4, 4 and 6, so reversal of the dominant omission does not imply complete preservation of $A$. These outcomes kept the first-skill distance, head and transplant analyses on digits, with the later language skill examined in Appendix~\ref{sec:app-nb}. A \emph{single-slot execution} account, in which filler-only training leads each skill to treat other blocks as no-ops, fits later-skill loss, the failed cues and reliable separate execution, but not partial loss of $A$. It remains unresolved, because the knockouts in Appendix~\ref{sec:app-nb} match none of its specified signatures.

\section{RQ2: Layout Robustness and Compatibility}
\label{sec:app-diag}

\begin{figure*}[t]
\centering
\includegraphics[width=0.75\textwidth]{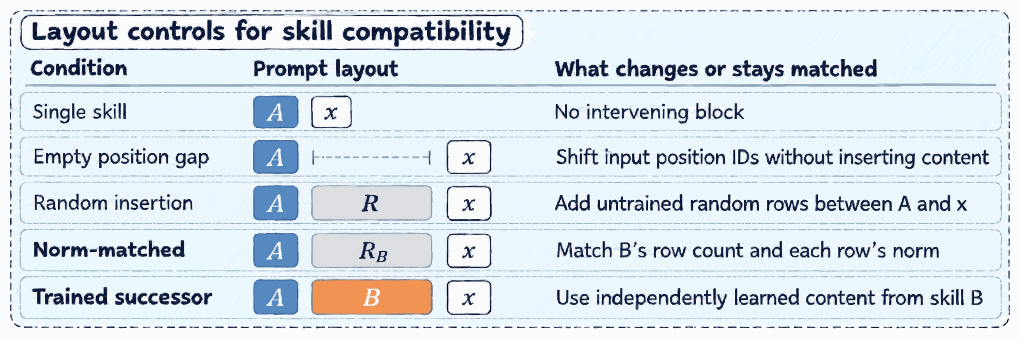}
\caption{\textbf{Layout controls change different properties of the prompt.} A protocol schematic, not measured activations, with the earlier skill and its input fixed within each comparison.}
\label{fig:layout-controls-protocol}
\end{figure*}

This appendix supports Sections~\ref{sec:diag-layout} and~\ref{sec:diag-compat}. Figure~\ref{fig:layout-controls-protocol} separates what a trained successor changes at once: random insertion adds content and positions, an empty gap changes only position IDs, and original and layout-augmented skills are separately trained populations.

\subsection{How Layout Changes Suppress an Isolated Skill}

\label{sec:app-d1}

Can an isolated skill fail when no second learned operation is requested? For each earlier skill $A$, we insert $k$ random embedding rows, 20 rows equal to a skill's starting vector (\sk{INIT}), $k$ copies of \sk{ID} or `the', \sk{REVERSE}, or a trained successor $B$. A complementary intervention inserts no material and instead increases the position IDs after $A$ by $k$, up to 40 positions. This artificial gap tests positional sensitivity without introducing successor activations. It is not the ordinary layout of a composed prompt.

Each condition uses the same 300 length-8 strings and masked decoding. Comparisons with $A$ alone use McNemar's test, with Holm correction over 20 controls within a unit. At the specified reference length $k=20$, presence below 20\% counts as loss and presence at or above 80\% as preservation. These cut-offs classify observable responses, rather than identify an internal cause.

Figure~\ref{fig:successor} summarises these conditions at twenty positions (left) and by span length (right), with medians over constituent units, or over pair-seed cells for trained successors. Its last left column gives the augmented 0.5B skills of Appendix~\ref{sec:app-l1}, and its dark grey dotted curve gives them with random rows. Filler-copy counts and position counts differ as detailed in Appendix~\ref{sec:app-setup}.

\input{tables/successor-distance_float.tex}

All 66 units keep $A$ at 80\% or above alone. A gap of 20 removes it in 63 units, with all three exceptions on Gemma, twenty random rows in 55, and random rows, \sk{INIT}, `the' and the gap all together in only 32. Thus the existence of layout sensitivity is more consistent than the response to any particular inserted material.

Short pretrained-operation spans are less disruptive. Five copies of \sk{ID} preserve median $A$ presence of at least 84.0\% on every base, and \sk{REVERSE} preserves at least 99.7\%. Twenty \sk{ID} copies reduce the median to at most 50.5\%. With a trained successor, $A$ falls below 20\% in 74 of 81, 73 of 81, 51 of 54 (seeds 2 and 3) and 75 of 81 pair-seed cells on 0.5B, 1.5B, 7B and Gemma, respectively. The corresponding pair counts, requiring a majority of seeds, are 25, 24, 24 and 26 of 27. Since presence includes outputs that omit only $B$, it can exceed composition accuracy: on 1.5B, \sk{IP}\sk{SL} has presence 29.0\% but exact composition accuracy 1.4\%.

The logged signatures, evaluated by unit majority on each base, separate loss after random rows only, loss after a gap while both \sk{ID} and `the' preserve $A$, and loss after a gap without both fillers preserving it. The third, broad distance-sensitive signature is the majority outcome on every base, in 13 of 18 units on Qwen2.5-0.5B, 15 of 18 on 1.5B, 9 of 12 on 7B and 15 of 18 on Gemma-3-1B. The second, narrower gap-sensitive signature occurs in 5, 3, 3 and 0 units, 7 of them \sk{INVERT\_POLARITY}. The remaining three Gemma units, also \sk{INVERT\_POLARITY}, keep 34.7\%, 74.0\% and 36.0\% after the gap and meet neither classification. No unit loses $A$ only after random rows.

\subsection{Does Neutral-Insertion Training Restore Composition?}

\label{sec:app-l1}

If unfamiliar neutral spans were sufficient to explain composition failure, training skills to tolerate such spans should improve their subsequent compatibility. We therefore retrain each skill independently with distractors, without composed labels. With probability 0.5, a training prompt receives one distractor at a random boundary before the digits while retaining its original target. The distractor is drawn with equal probability from a frozen 20-row random block, a 5- or 10-row random block, or 1 to 20 copies of `the'. The remaining recipe and three skill seeds match Appendix~\ref{sec:app-setup}.

Composition is summarised by the median over the 27 pairwise seed medians. The specified benchmark is 69\%, the tuned SPoT result with eight labels on 0.5B, together with at least five skills reaching 90\% alone. Partial success begins at 10\%. These thresholds were carried unchanged to Qwen2.5-1.5B and Gemma-3-1B before their runs, and to 7B seeds 2 and 3 after the smaller-base results but before the 7B runs.

On Qwen2.5-0.5B, all constituents achieve 100\% alone and all 18 units preserve $A$ at 80\% or above after 20 random rows and 20 copies of `the'. Yet median composition accuracy is 0.3\% (mean 3.6\%), below the partial-success threshold of 10\%, with no pair reaching 69\% (Table~\ref{tab:app-l1}). An empty gap still removes $A$ in all 18 units, including 15 in which both \sk{ID} and `the' preserve it. A trained successor removes $A$ in 73 of 81 cells (26, 24 and 23 pairs in the respective seeds), compared with 74 of 81 before augmentation, and both sets have 25 of 27 pairs with a majority of affected seeds. Of the augmented-skill outputs, 90.6\% match none of $A(x)$, $B(x)$, $B(A(x))$ or $x$, and only 3.0\% equal $B(x)$. The residual failure usually extends beyond clean omission of $A$.

\input{tables/app-l1_float.tex}

The other bases show the same separation. On Qwen2.5-1.5B, all skills achieve 100\% alone and all 18 units preserve $A$ after random rows and `the', but composition has median 0\% and mean 0.5\%, with no pair reaching 10\%. Robustness is not uniform across untrained material: \sk{INIT} preserves only 5 of 18 units, the gap removes $A$ in 15 of 18, and a trained successor removes it in 68 of 81 cells and a majority of seeds for 23 of 27 pairs. Gemma-3-1B offers a stronger positional control. Its augmented skills all achieve 100\% alone, and a gap of 20 preserves $A$ at 80\% or above in 15 of 18 units, removing it below 20\% in none. No unit meets the earlier failure signatures, which require random rows or a gap to remove $A$, although 7 of 18 still decline significantly relative to $A$ alone after the gap (McNemar, Holm). Longer interventions remain damaging, with median presence 18.7\% after 40 \sk{ID} copies and 47.3\% after a gap of 40. Composition still has median 0\% and mean 4.3\%, and a trained successor removes $A$ in 70 of 81 cells and a majority of seeds for 23 of 27 pairs. This clearly separates neutral-layout robustness from successful composition, but it does not establish the norm-controlled distinction tested next. The 7B extension has unreliable constituents: only four of six skills reach 90\% alone, with \sk{MIRROR} in seed 2 and \sk{INVERT\_POLARITY} in seed 3 the two failed units among twelve. No pair reaches 10\%, including pairs composed of skills that work in both seeds. The other ten units preserve $A$ after 20 random rows and 20 copies of `the' but lose it after a gap of 20, and a trained successor removes $A$ in 50 of 54 cells, including 41 of the 45 cells whose first skill works alone.

Across the four bases, robustness to the tested neutral insertions therefore does not transfer automatically to a learned successor. The intervention does not establish two independent or additive causes, and it does not cover every possible augmentation recipe.

\subsection{Can Length and Norm Explain Successor~Harm?}

\label{sec:app-compat}

The augmentation result still permits a magnitude explanation: a trained successor may simply introduce larger embedding rows than the neutral distractors. For each directed pair, we replace $B$ with a random block of the same length and scale each row to the norm of the corresponding row of $B$. A fixed direction draw per successor retains row count and magnitude while changing directions and within-block structure.

The amended rule requires at least 18 of 27 pairs, by seed majority, to preserve $A$ at 80\% or above after the random block and lose it below 20\% after trained $B$. This rule was set for augmented skills with the insertion results, including the reused trained-$B$ scores, already known. No base meets the original extra requirement that every pair preserved under random rows lose $A$ under trained $B$. The two-thirds cell criterion provides a separate sensitivity check. These distinctions limit the confirmatory status of the comparison.

\begin{figure*}[!htbp]
\centering
\includegraphics[width=\textwidth]{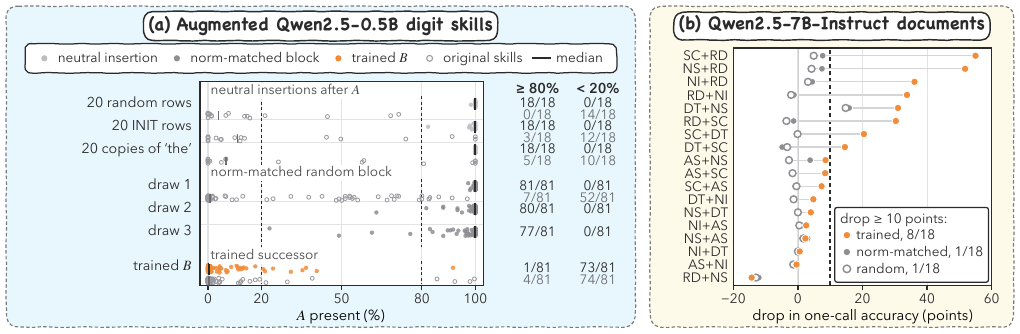}
\caption{\textbf{Trained blocks harm more than norm-matched random blocks on augmented 0.5B skills and, descriptively, on 7B document pairs.} (a) Augmented Qwen2.5-0.5B digit skills, original skills hollow, dashed lines marking the column-header count thresholds, and (b) one-call accuracy drop per Qwen2.5-7B-Instruct document pair, dashed at 10 points.}
\label{fig:compat-detail}
\end{figure*}

\paragraph{The contrast on augmented Qwen skills.}
For augmented Qwen2.5-0.5B skills, the random block preserves $A$ in all 81 cells, whereas trained $B$ removes it in 73. All 27 pairs preserve $A$ under the random block, 25 show the required contrast, and trained $B$ gives significantly lower presence in every pair and seed (McNemar, Holm over pairs). Two more direction draws yield 25 and 24 qualifying pairs, and the three draws retain presence in 81, 80 and 77 of 81 cells (Figure~\ref{fig:compat-detail}a, which also shows the original skills). The relative norm-matching error is at most 0.00035, and trained-row median norm is 3.647, compared with 0.592 for ordinary insertion-control rows. Qwen2.5-1.5B likewise preserves $A$ in all 81 random-block cells, loses it in 68 trained-$B$ cells, qualifies on 23 of 27 pairs and shows lower presence after trained $B$ in every pair and seed. Both Qwen bases pass the cell criterion, with 73 and 68 of 81 cells against the required 54. On these bases, successor length and per-token norm alone are insufficient to explain the residual~\mbox{interference}.

\paragraph{Why the contrast does not generalise.}
On Gemma-3-1B, only 9 of 27 pairs qualify. The random block preserves $A$ in 38 of 81 cells and removes it in 16, and by seed majority it preserves 11 pairs and removes 4, all beginning with \sk{SWAP\_PAIRS}. The cell contrast holds in 31 of 81, below 54. For 7B, the majority rule requires both seeds 2 and 3 to qualify. Only 12 of 27 pairs do so, the random block removes $A$ in five pairs, all beginning with \sk{MIRROR}, and the cell criterion fails with 32 of 54 against the required 36. The two failed constituent units have presence at most 1.0\% after either block in their nine cells, and the specified rule contains no acceptance filter. On these two bases, the matched block itself fails to preserve $A$ in 16 of the 18 and 13 of the 15 pairs that do not qualify. In a post hoc test, $A$ present is still significantly lower after trained $B$ in every seed for 18 and 17 of 27 pairs, rather than all 27. Thus norm-matched random blocks themselves remain harmful on these bases.

A post hoc restriction to working first skills leaves 45 of the 7B cells. The random block preserves 36 of them and removes 7, five \sk{MIRROR} and two \sk{SHIFT\_RIGHT} cells, all in seed 3, and the required contrast occurs in 32 of 45, exceeding two thirds. Re-evaluating each pair only on seeds whose first skill works nevertheless yields 14 of 27 qualifying pairs, still below 18. The same two seed-3 skills also lose $A$ after 20 \sk{INIT} rows and 20 \sk{ID} copies, a pattern noticed after the runs. Acceptance failure explains part, but not all, of the 7B shortfall.

For original, non-augmented skills, the random block itself removes $A$: in 52 of 81 cells on 0.5B and by seed majority in 24 and 26 of 27 pairs on 1.5B and Gemma. These descriptive results are compatible with the earlier layout sensitivity. The bounded positive finding is therefore a norm-controlled separation on \emph{augmented} Qwen2.5-0.5B and 1.5B. Learned directions, activation patterns and within-block structure remain possible contributors, and the experiment does not identify semantic conflict as the cause. Figure~\ref{fig:compat-detail}b plots the document-prompt contrast of Appendix~\ref{sec:app-document-blocks} on Qwen2.5-7B-Instruct, one point per different-field pair. Learned blocks affect 8 of 18 pairs against one for norm-matched blocks, below the original twelve-pair threshold, and this pair-class split was selected after the 7B results. Pair labels join the two operations with a plus sign: NS is name swap, NI name initial, SC surname caps, AS add shipping, DT double total and RD round down.

\subsection{Sensitivity to Block Length and Skill Identity}

\label{sec:app-ll}

The preceding comparisons use 20-row skills. To test whether this choice fixes the observed fragility, we retrain all six Qwen2.5-0.5B digit skills with 5 and 40 rows using seed 1 and the same recipe. Table~\ref{tab:app-ll} shows that every skill achieves 100\% alone, but a gap of 20 and 20 random rows both remove $A$ in 5 of 6 skills at 5 rows and 4 of 6 at 40 rows, compared with 6 of 6 at 20 rows. For 40-row skills, the gap leaves presence at most 0.7\% in every skill, below the proposed robustness threshold. A trained successor removes $A$ in 27, 26 and 25 of 27 pairs at 5, 20 and 40 rows, and median joint accuracy is 0\% at each length. Per-skill presence values are reproduced by the released code. Within this single-base, single-seed comparison, block length neither restores composition nor moves the specified layout threshold.

\input{tables/app-ll_float.tex}

A post hoc pattern distinguishes \sk{INVERT\_POLARITY}, the complement of the pretrained \sk{POLARITY}, from the digit-moving skills. Twenty \sk{ID} copies preserve it at 80\% or above in 10 of 11 units, compared with 2 of 33 units for the three digit-moving skills, and every pair retaining its earlier skill at 20\% or above after a trained successor begins with \sk{INVERT\_POLARITY}, except \sk{MIRROR}\sk{IP} on 0.5B. The cause of this skill-specific robustness remains open. The two original 0.5B pairs whose seed medians exceed 69\%, \sk{MIRROR}\sk{IP} and \sk{IP}\sk{SR}, fall from 97.0\% and 88.3\% to 0\% and 13.3\% after augmentation. Because these pairs were selected for high original performance, the comparison is descriptive and susceptible to regression towards the mean.

\subsection{Is a Trained Successor Harmful Only Because It Is Unfamiliar?}
\label{sec:app-neo-reverse}

A trained successor could suppress the earlier skill because it computes a function the skill never met next to it in training, or because the skill never met that block at all. To hold the function fixed, we use \sk{NEO\_REVERSE}: 20 new rows trained alone on Qwen2.5-1.5B with the recipe of Appendix~\ref{sec:app-setup} to compute the pretrained \sk{REVERSE}, one block per skill seed. Table~\ref{tab:app-neo-reverse} compares it with the pretrained \sk{REVERSE} next to each original, non-augmented digit skill. Its upper panel places each block after the skill and its lower panel places it first. This is a post hoc, descriptive reading of existing runs.

\input{tables/app-neo-reverse_float.tex}

\paragraph{The learned block alone.}
\sk{NEO\_REVERSE} scores 100\% alone on the 2000 test strings in every seed and at least 99.5\% at lengths 4 to 8. It fails held-out length 9 (12.5\%, 6.0\% and 11.5\%), as the base's own \sk{REVERSE} does (9.0\%). It therefore fails the original acceptance rule, with held-out means of 70.8\%, 68.7\% and 70.3\%, and passes only the revised one (Appendix~\ref{sec:app-acceptance-scope}). Next to each of the seven pretrained operations, in either order, it composes in one call at 98.0\% to 100\%, including \sk{ADD}, which was held out of its training.

\paragraph{The learned block after a skill.}
On the same 2000 length-8 strings with masked decoding, $A$ present is 96.9\% to 100\% after the pretrained \sk{REVERSE} in all 18 skill and seed cells. After \sk{NEO\_REVERSE}, it is below 20\% in 16 of them, at most 5.7\% for the five skills other than \sk{INVERT\_POLARITY}, and lower in every cell (exact McNemar, $p<10^{-60}$ each). \sk{INVERT\_POLARITY} is the exception, with $A$ present at 18.0\%, 89.4\% and 87.5\% and exact match at 9.0\%, 5.0\% and 87.3\% in seeds 1 to 3. In seed 2, 84.5\% of outputs equal $A(x)$, so $A$ survives there mostly because the reversal is not applied, and only seed 3 composes.

\paragraph{The learned block before a skill.}
With \sk{NEO\_REVERSE} first, one-call exact match is at most 4.0\% for the five skills other than \sk{INVERT\_POLARITY} and 26.0\% to 81.2\% for \sk{INVERT\_POLARITY}, against 98.0\% to 100\% with the pretrained \sk{REVERSE} first. The \sk{REVERSE}-first values come from the 100-item acceptance evaluation, not from the same strings, so this direction is descriptive only.

\paragraph{Reading.}
Matching the successor's function does not remove the interference in this setting, but the comparison does not separate learned rows from unfamiliarity: the pretrained \sk{REVERSE} was a training neighbour of every skill, while \sk{NEO\_REVERSE} and the digit skills never met in training. The result is consistent with harm from a block the skill never saw in training, and equally with harm from learned rows as such. It covers one function, one base and three seeds, and the learned block passes only the revised acceptance rule. Appendix~\ref{sec:app-l2} tests the familiarity reading.

\subsection{Does Training with Learned Partners Reach Unseen Partners?}
\label{sec:app-l2}

Layout-augmented skills see random rows and neutral text but never another learned skill, so the trained successor of Section~\ref{sec:diag-compat} is out of their training distribution. We test whether training next to learned partners closes the gap for partners the skill never saw. The design, splits and readout were logged internally before these runs (Appendix~\ref{sec:app-setup}).

\paragraph{Recipe.}
Each digit skill is retrained with the layout-augmented recipe of Appendix~\ref{sec:app-l1}, except that in two-operation training samples the neighbouring operation is, with probability 0.5, a frozen learned skill instead of a pretrained one. The partner block holds the 20 original rows of another core skill, with identity and seed drawn uniformly, placed before or after the skill with equal probability, and the target is the composed function in that order. Partner rows receive no gradient, and validation samples contain no partner.

\paragraph{Held-out partners.}
Each skill's five possible partners are split into groups of two and three. One run trains with the first group and holds out the second, and a second run does the reverse, so each of the 27 directed pairs is held out once and seen once. Evaluation uses the digit strings of Appendix~\ref{sec:app-d1}, the same-seed original partner skills, the one-call condition and the norm-matched block of Appendix~\ref{sec:app-compat}.

\paragraph{Readout.}
Cells whose augmented skill scores below 90\% alone are excluded and reported. On held-out pairs, composition is the median over pairs of the per-pair seed median of one-call accuracy, against the benchmark of Appendix~\ref{sec:app-l1}, and compatibility counts pairs whose seed majority keeps $A$ present at 80\% or above after the partner, against the bar of 18 of 27. Generalisation requires both. If only seen pairs pass both, the gain is specific to the partners used in training, and if seen pairs pass neither, training next to learned partners does not repair composition under this recipe.

\paragraph{Result.}
Table~\ref{tab:app-l2} reports composition and pairs keeping $A$ over the 27 directed pairs with masked decoding, where seen and held-out say whether the partner's identity was used in the skill's training. Its upper panel is the pre-specified evaluation, and its lower panel is the post hoc, descriptive check with new partner rows below, whose medians are also taken over partner seeds. On both bases, all 36 retrained skills score at least 90\% alone, so no cell is excluded, and the pre-specified reading is that the gain is specific to the partners used in training. With held-out partners, the composition median is 0\% on Qwen2.5-0.5B and on 1.5B, and $A$ is kept in 2 and 1 of 27 pairs, against the benchmark of 69\% and the bar of 18. With seen partners, the medians are 98.0\% and 95.3\%, with $A$ kept in 21 and 19 pairs. After the norm-matched block, $A$ present has median 100\% in the held-out cells of both bases. Against the layout-augmented skills on the same strings, $A$ present after a held-out partner is significantly higher in 11 to 13 of 27 pairs per seed on 0.5B and 12 to 18 on 1.5B, and lower in at most four (exact McNemar tests with Holm correction, descriptive). Median one-call accuracy is at most 0.7\% under both recipes.

\input{tables/app-l2_float.tex}

\paragraph{New rows of seen partners.}
The seen pairs of the upper panel reuse the exact partner rows placed next to the skill in training, so they also reward fitting those rows. As a post hoc, descriptive check, we evaluate every retrained skill against the original skills of seeds 104 and 105, which training never used, so each partner identity comes with new rows. Every such partner scores at least 90\% alone, so no cell is excluded. With new rows of seen identities, the composition median falls to 12.7\% on 0.5B and 0.5\% on 1.5B, and $A$ is kept in 3 and 1 pairs. With new rows of held-out identities, the medians are 0\% and 0\%, with $A$ kept in 2 and 1 pairs. Under the tolerance of 10 points fixed before these runs, 1.5B reads as specific to the trained rows. 0.5B reads as in between, although its value lies far nearer the held-out level than the same-row level.

\paragraph{Reading.}
Training next to learned partners yields composition with the partner rows used in training, but not with held-out partner identities and largely not with new rows of the seen identities. Under this recipe, the result is consistent with the gained compatibility attaching mainly to particular partner rows rather than to a partner's function. Together with the pretrained operations that digit skills met in training (Appendix~\ref{sec:app-acceptance-scope}) and the learned block computing \sk{REVERSE} (Appendix~\ref{sec:app-neo-reverse}), this is consistent with compatibility forming mainly between blocks that met in training, although the learned block also composes with \sk{ADD}, which it never met in training. The evidence covers only two Qwen bases, three skill seeds, two partner seeds and a single~recipe.

\section{RQ2: Skill Documents and Textual Support}
\label{sec:app-c1}

This appendix supports Section~\ref{sec:diag-carrier} (RQ2). Documents also add information, training history and prompt length, so the controls below constrain a representation account without isolating representation as a single cause.

\subsection{Where Documents Improve Composition}

Before the language runs, removal of interference was defined as equivalence to two calls on at least 80\% of eligible pairs with none still interfering, and persistence as at least 50\% still interfering. Eligible pairs are interfering soft-skill pairs whose document constituents also pass the single-skill gate, which excludes 10 and 5 of 18 skill-document draws on Qwen2.5-0.5B-Instruct and 1.5B-Instruct.

Documents interfere on all four eligible pairs at 0.5B and nine of twelve at 1.5B, with two equivalent pairs on the latter base. At 7B, 17 of 29 eligible pairs are equivalent and ten interfere. One of these is a same-name pair on which documents score 0\% both in two calls and in one call, so the remaining 16 matches are the different-field pairs. Thus the full comparison meets the persistence rule on the two smaller bases and neither rule on 7B. No base satisfies the specified removal rule (7B values in Table~\ref{tab:carrier}, means over three skill seeds or document draws).

\IfFileExists{tables/carrier_float.tex}{

\begin{table}[!t]
\centering
\footnotesize
\setlength{\tabcolsep}{2pt}
\hlclip{

\begin{tabular}{@{}lcccc@{}}
\toprule
\hlheadrow  & \multicolumn{2}{c}{\hlhd{Gap}} & \hlhd{Soft} & \hlhd{Text} \\
\cmidrule(lr){2-3}
\hlheadrow \hlhd{Pair class} & \hlhd{Soft} & \hlhd{Text} & \hlhd{int.} & \hlhd{eq./int.} \\
\midrule
Different fields (18) & \hlkey 97.8 & \hlkey \hlbest{3.8} & 18/18 & 16/1 \\
Same name field (6) & 91.8 & \hlbest{9.6} & 6/6 & 1/3 \\
Same total (6) & 99.9 & \hlbest{62.7} & 6/6 & 0/6 \\
All (30) & 97.0 & \hlbest{16.7} & \hlfail 30/30 & \hlfail 17/10 \\
One call (30) & 2.0 & \hlbest{70.2} & & \\
Two calls (30) & \hlbest{99.0} & 86.9 & & \\
\midrule
\hlheadrow \hlhd{Condition} & \hlhd{Single} & \hlhd{One call} & \multicolumn{2}{c}{\hlhd{One call, D pairs}} \\
\midrule
Text, on demand & \hlbest{97.6} & \hlbest{69.6} & \multicolumn{2}{c}{\hlbest{89.4/92.6/93.1}} \\
Text, all six loaded & 84.6 & 39.2 & \multicolumn{2}{c}{40.6/50.8/62.1} \\

\bottomrule
\end{tabular}
}
\caption{\textbf{Documents help 7B composition most on different-field pairs and weaken when all six are loaded.} With accuracies in \% and gaps in points, coral marks interference or a missed removal criterion, yellow values quoted in the main text, and bold the better value of each compared pair.}
\label{tab:carrier}
\end{table}
}{}

\input{tables/app-carrier_float.tex}

The split by pair class was chosen after observing the 7B results. Figure~\ref{fig:app-carrier} shows that documents are equivalent to two calls on 16 of 18 different-field pairs but interfere on all six non-commuting total pairs. Even this advantage depends on which documents are loaded: loading all six documents reduces different-field one-call accuracy to 40.6\%, 50.8\% and 62.1\% across draws, compared with 89.4\%, 92.6\% and 93.1\% when only invoked documents are loaded.

In Figure~\ref{fig:app-carrier}, markers average three skill seeds or document draws and bars keep the per-seed 90\% intervals. An open square marks an undetermined label, a cross an ineligible pair and a grey tick a pair not eligible for documents, and the last two show no value. Class means above each panel use only the pairs that show a value, so the 7B same-name document mean is $-5.1$ points there and 9.6 points over all six pairs in Figure~\ref{fig:fig_carrier}. Across 0.5B, 1.5B and 7B, soft blocks interfere on 20 of 20, 20 of 20 and 30 of 30 eligible pairs. At 7B, the open square near $-100$ points marks a pair on which documents fail in two calls but succeed in one.

\subsection{Examples and Constituent Competence}
\label{sec:app-document-examples}

Worked examples supply both an operation specification and an output format. To test whether they carry the document advantage, three specified 7B conditions retain only document procedures, only worked examples with skill names, or soft blocks with the same two examples per skill. The first two panels of Table~\ref{tab:app-c1} give means over three document draws or skill seeds, with one-call accuracy over all 30 pairs and gaps by pair class, each gap taken against the stored full-document or soft two-call reference of the same draw or seed. Procedure-only documents are weaker individually, with single accuracy 80.3\% rather than 97.7\%, and their gap on different-field pairs is 46.4 points. Only two pairs remain within 10 points of the full-document two-call reference in every draw, short of the specified twelve. Conversely, adding examples to soft blocks leaves only five pairs below 10\% one-call accuracy, fewer than the predicted fifteen. With examples matched, the two-call-minus-one-call gap remains 13.1 points for example-based documents and 71.5 points for soft blocks.

\input{tables/app-c1_float.tex}

Mean different-field one-call accuracy is 49.1\% for procedure-only documents, 82.4\% for examples only and 27.5\% for soft blocks with examples, against 91.7\% for full documents and 1.2\% for soft blocks, and the released code reproduces the per-pair values behind these means. Worked examples therefore contribute to the document advantage and partly improve the soft representation. Removing them changes constituent competence as well as composition, so the two comparisons do not isolate a composition-specific contribution. The example-based documents still include descriptive skill names, so the matched comparison does not isolate examples as the sole difference.

\subsection{Controlling Information and Context}
\label{sec:app-c2}

The remaining controls ask whether concise instructions suffice, and whether adding learned rows disrupts a working document prompt. They evaluate all 30 directed pairs on Qwen2.5-7B-Instruct with new output IDs masked. Each of three units pairs one document draw with one skill seed. Pair values are means over these units, rounded before thresholding, and the specified decisions use the 18 different-field pairs. Shared conditions reproduce the stored primary results, and gaps use the document two-call reference for document-based rows and the soft one otherwise (last two panels of Table~\ref{tab:app-c1}).

\paragraph{Short descriptions, names and order cues.}
Replacing each soft block with 20 tokens of terse skill description fails to reproduce the document result. Fourteen of eighteen pairs remain below 10\% one-call accuracy, with a different-field mean of 7.9\%, and none is within 10 points of two soft calls, satisfying the specified failure rule. However, the terse descriptions of \sk{NL\_NAME\_INITIAL} and \sk{NL\_SURNAME\_CAPS} score 76.0\% and 7.7\% alone and occur in eleven of the fourteen failing pairs. Among the six pairs without these descriptions, three remain below 10\% and five below 20\%. Much of this control's failure therefore reflects unreliable constituent instructions, rather than a clean test of compositionality at matched length.

Adding short skill names before the soft blocks and joining them with ``, then'' brings none of the eighteen pairs within 10 points of two soft calls. Nine remain below 10\%. Neither aggregate count reaches the twelve-pair decision threshold, although the separate unit counts below 10\% are fourteen, thirteen and fourteen. Names improve mean different-field one-call accuracy from 1.2\% to 13.0\%, still below 91.7\% for documents, while single-skill accuracy falls from 99.5\% to 79.9\%. Names without soft blocks compose on no pair. The tested names and order cue do not restore the document-level performance, although they are not behaviourally inert.

A secondary plain-sentence carrier, evaluated on one draw, composes poorly mainly because only about half of outputs parse as records. It reaches 18.3\% one-call accuracy over all pairs and 10.0\% on different-field pairs, where 53.9\% of outputs parse, with 90.2\% single-skill accuracy. Inserting ``then'' or \sk{ID} between soft blocks gives little improvement, 4.7\% and 2.7\% one-call accuracy over all pairs against 2.0\% for plain soft blocks, whose single-skill accuracy is 99.5\%.

\paragraph{Learned blocks at matched length and norm.}
\label{sec:app-document-blocks}
Holding the document fixed, adding its two invoked soft blocks at the end of the Steps line reduces one-call accuracy by at least 10 points on 8 of 18 different-field pairs. Adding two blocks of 20 random rows does so on one pair. This contrast is descriptive: neither reaches the specified twelve-pair threshold for widespread disruption, and failure to cross the threshold does not, by itself, establish missing textual information as the alternative cause. A neutral-token block at the same positions was not tested. Figure~\ref{fig:compat-detail}b plots these per-pair drops, so it shows the same observations rather than an independent replication.

To ask whether embedding magnitude reproduces this contrast, we replace the added trained blocks with random blocks matching their row counts and corresponding L2 norms, with directions drawn once per slot. The thresholds were set after observing trained-block results, when ordinary random blocks had already affected one pair: a drop of at least 10 points on eight or more pairs would reproduce the observed harm, whereas two or fewer affected pairs would count as sparing. Norm-matched blocks affect one of eighteen different-field pairs, the same pair affected by ordinary random blocks. Two further specified direction draws, for which eight affected pairs would have withdrawn the sparing result, again affect only that pair. Across the three draws, relative norm errors are at most $1.9\times10^{-4}$ and mean absolute cosine similarity to the trained rows is at most 0.014. Length and per-token norm alone therefore do not reproduce the additional harm from learned blocks in this subset, within the post hoc pair-class selection and thresholds and the unmet twelve-pair disruption threshold. Learned directions or their induced activations remain candidate explanations.

Outside the specified decision set, norm-matched blocks affect four of thirty pairs compared with twelve for trained blocks, yet lower mean same-field one-call accuracy more, by 8.2 versus 1.6 points. They are therefore not universally benign, and the different-field contrast should not be generalised to all operations or prompt~\mbox{contexts}.

\section{RQ2: Internal Interventions and Their~Limits}
\label{sec:app-where}

This appendix supports Section~\ref{sec:diag-probes} (RQ2): the tested interventions did not identify a local mechanism sufficient to explain the interference.

Table~\ref{tab:transplant} maps each tested digit account to its outcome: whether a working state restores the missing behaviour, and whether one site, a few heads or a layer band explains the loss. Counts are descriptive over shared pair-seed cells or constituent units. Transplant rows follow exploratory rules set after initial runs, common-site rows are later reanalyses that are not localisation evidence, and no digit probe meets its specified base-level criterion. Per-cell restoration (19 of 30 smaller-base cells, generation accuracy) and the common-site count (7 of 39 cells, first-digit logit difference) use different metrics, so they are not successive success rates. Successful restoration shows what the selected intervention can do, not the pathway of natural failure, and an unsuccessful search rules out only the tested interventions, without establishing a distributed mechanism. None of these internal results is a premise for the composition, layout or retention findings in Sections~\ref{sec:revisit}, \ref{sec:diag-layout}--\ref{sec:diag-carrier} and~\ref{sec:repair}.

\IfFileExists{tables/transplant_float.tex}{

\begin{table*}[!htbp]
\centering
\footnotesize
\setlength{\tabcolsep}{5pt}
\hlclip{

\begin{tabular}{@{}lccccc@{}}
\toprule
\hlheadrow \hlhd{Test} & \hlhd{0.5B} & \hlhd{1.5B} & \hlhd{7B} & \hlhd{Gemma} & \hlhd{Bar} \\
\midrule
Transplant, per cell & 7/12 & 6/10 & 0/2 & 6/8 & -- \\
\quad specific & 7/7 & 5/6 & -- & 6/6 & -- \\
\quad 7B single-site search & -- & -- & \hlkey 1/8 & -- & 4/8 \\
Fixed site per base & 7/12 & 4/10 & 5/9 & 1/8 & -- \\
\quad global null & 3/12 & 1/10 & 3/9 & 0/8 & -- \\
\addlinespace[2pt]
\hlctrlrow Full-depth knockout, distance & 12/12 & 10/10 & -- & -- & -- \\
\hlctrlrow \quad content & 0/12 & 0/10 & -- & -- & 6/9, 6/8 \\
Early-layer knockout, restores & \hlfail 0/12 & \hlfail 0/10 & -- & -- & 4/7, 3/5 \\
\addlinespace[2pt]
Head sets, units with $k^*{\le}16$ & \hlfail 3/9 & \hlfail 2/9 & \hlfail 0/3 & -- & 2/3 \\
\quad attention mass at most half wherever $A$ is lost & \hlfail no & \hlfail no & \hlfail no & -- & ratio 0.5 \\
\quad gain from their outputs & \hlfail 0.2 & \hlfail 0.8 & \hlfail 0.4 & -- & $+$10 \\
Layer-band transplant, band restores & -- & -- & \hlkey 1/5 & -- & 3/5 \\
\bottomrule
\end{tabular}
}
\caption{\textbf{Internal interventions restore some cells, but no tested site, head set or layer band explains the loss.} With gains in points, teal meets the bar and coral misses it, yellow marks main-text values that also miss it and gray rows are the gap control.}
\label{tab:transplant}
\end{table*}
}{}

\subsection{When Does a Preserved State Restore Composition?}

Activation replacement asks whether the failing computation can use a state from a context that retains $A$. A cell is a pair-seed combination in which one-call execution loses the first skill. At one layer, we replace the residual stream at the input digits, the `=' token, or both. These positions follow the successor block. This tests recoverability at a selected site. Identifying a shared site requires the separate sensitivity analysis below.

\paragraph{Sources and controls.}
The source is \texttt{[A][ID][B]x} when its $A$ presence is at least 50\%, a working context with a pretrained successor when presence is at least 90\%, or the single-skill context \texttt{[A]x}. Source, site and layer are selected by logit-difference gain on the selection pool, subject to the stored direct-share criterion below 0.5. Generation is then evaluated on a disjoint report pool. The intervention follows a single-site pure-indirect-effect construction~\citep{vanderweele2014unification}, rather than attribution patching~\citep{kramar2024atp}.

Success requires a composition exact-match gain of at least 10 points with exact McNemar $p<0.05$. To distinguish the intended source from a generic replacement effect, specificity additionally requires a margin of at least 10 points over a wrong-skill source in which another new skill replaces $A$. These selection and interpretation rules were fixed after initial runs, so the analysis is exploratory.

\paragraph{Restoration depends on the cell and the source.}
The true-source patch succeeds in 7 of 12 cells on Qwen2.5-0.5B, 6 of 10 on 1.5B, 6 of 8 on Gemma-3-1B and none of two on 7B (Table~\ref{tab:transplant}), and failed cells improve by at most 7.2 points. Of the 19 smaller-base successes, 18 are specific. The exception is \sk{SL}\sk{IP} in 1.5B seed 2, where the wrong-skill source produces higher composition accuracy than the true source, 20.4\% versus 11.4\%. The released code reproduces every per-cell exact match and common-site value.

These successes do not all test the same source content. Seven of nineteen use \texttt{[A]x}, which computes only the earlier operation: five of seven successes on 0.5B, none of six on 1.5B and two of six on Gemma. The other twelve use a working pretrained-successor context whose answer determines the composed answer. For $B=\sk{IP}$, for example, \texttt{[A][POLARITY]x} produces the digit-wise complement of the desired composition. The wrong-skill comparison tests source specificity, but does not remove this close relationship between some source answers and the target answer.

\paragraph{Recoverability does not establish necessity.}
The complementary necessity tests are negative. Mean ablation at `=' of the top $k^*$, $4k^*$, 64 or 256 readout coordinates of the earlier skill alone leaves its accuracy unchanged in all 32 cells (12, 10, 8 and 2 per base). Here $k^*$ is the minimum number of residual-stream coordinates at `=' that a sparse linear probe needs to reach a fixed fraction of the best probe's AUROC, aggregated by the median over decisions and seeds. This $k^*$ is a property of the measured readout and differs from the head count in Appendix~\ref{sec:app-ha}. Steering along execution, success or probe-value directions succeeds in no cell. The positive result is therefore the existence of recoverable state under selected replacements. Neither the ablations nor steering identify coordinates whose normal contribution explains the loss.

\subsection{Do Successful Patches Share a Location?}

Per-cell selection can establish recoverability while choosing a different intervention for almost every failure. To ask whether that evidence extends to a common location, a later reanalysis chooses one source, site and layer per base by mean normalised logit-difference gain on the selection pool. It selects the digit site at layer 15 on every base. The evaluation covers the 39 cells of the main families, including nine from the wider 7B search below. The two cells of the earlier 7B block are reported separately (0 of 2).

The apparent support depends on how broadly the controls are compared. At the common configuration, logit-difference gain exceeds the largest same-position control in 17 of 39 cells, but exceeds the largest control at any site or layer in only seven (Table~\ref{tab:transplant}). This post hoc reanalysis evaluates each common configuration on 500 report items. For the gold first digit's top-1 rate, it exceeds every same-position control in 15 of 39 cells, with median margin of $-1.0$ points. The original per-cell choice agrees with the common configuration in only 2 of 12 cells on 0.5B, none of ten on 1.5B, two of eight on Gemma and none of nine on 7B.

Generation accuracy was neither rerun nor stored at the common site, so these proxy comparisons cannot establish a common generative rescue. The two analyses support recoverable states in individual cells, sensitive to where each cell is patched, without locating a shared site.

\subsection{Does Access to the Successor Explain the Loss?}
\label{sec:app-heads}

A successor could alter the first skill's contribution through the attention it receives. Removing access to its keys appears to test that account directly, but also changes the effective layout. This consequence must be established before interpreting a knockout as pathway evidence.

\remarkpar{rem:shift}If all positions after a block are prevented from reading that block's keys at every layer, the remaining positions retain their original position IDs. Removing access to $B$ in $[A][B]x$ therefore computes the corresponding $[A]x$ context with the input shifted by the length of $B$. It implements the position-gap control, rather than returning to the original single-skill layout.\par

\paragraph{Full-depth removal is a layout control.}
The specified study uses Qwen2.5-0.5B and 1.5B, excluding Gemma because of its QK normalisation and BOS attention sink. A knockout inserts the most negative finite value into the additive attention mask before softmax, preventing selected queries from reading selected keys~\citep{geva2023dissecting}. Teacher-forced exact match is used after requiring at least 98\% agreement with greedy decoding. Configuration selection uses 300 items and reporting uses 500 others. The primary set contains seven specific transplant successes on 0.5B and five on 1.5B. Runs use bfloat16 and float32, respectively, following the precision check in Appendix~\ref{sec:app-setup}.

The original restoration threshold required six of nine \sk{IP}-ending cells on 0.5B and six of eight on 1.5B to recover $A$ when digits, `=' and answer inputs could not read successor keys at any layer. Before these runs, Remark~\ref{rem:shift} instead motivated interpreting the intervention as an implementation check against the gap control. It leaves presence at 0\% in all twelve 0.5B cells and at most 3.8\% in all ten 1.5B cells, so $A$ stays lost in all 22 digit cells, while $A$ alone remains at 100\%. A positive control blocks access to $A$'s own block in \texttt{[A]x}, reducing presence from 100\% to 0\% in every cell. The implementation therefore removes the intended access, but the full-depth result supplies no independent explanation beyond the gap control.

\paragraph{Restricting the intervention does not recover a selective effect.}
Limiting the knockout to early layers avoids eliminating the successor throughout the network. A cutoff is selected on the selection pool. Success requires report-pool gain of at least 10 points with Holm-adjusted $p<0.05$, a margin of 10 points over late-band, mean-ablation and equal-norm controls, and a larger improvement for the composed answer than for alternative outputs. These comparisons distinguish selective restoration from a generic perturbation gain. No cell succeeds among twelve on 0.5B or ten on 1.5B (0 of 22), below the respective thresholds of four of seven and three of five primary cells. Neither the composed answer nor $A(x)$ returns in 38 of 44 query-set tests. The other six gain at least 10 points but fail the control or multiplicity requirements.

\paragraph{Carrying a patch's gain is a different causal question.}
We separately decompose the original transplant gain into attention and MLP outputs. Attention alone carries at least half the gain in five of seven primary 0.5B cells, with shares 0.60 to 1.62 and MLP shares at most 0.64, and in two of five 1.5B cells, where MLP leads in one. In the remaining two cells per base, neither component alone restores anything. The literal decision rule would label these zero-zero cases MLP-led, giving a three-of-five MLP majority on 1.5B. We report them as unclassified, a post hoc interpretation that removes this majority. Attention thus carries most of the selected restoration effect on 0.5B, and no dominant component is assigned on 1.5B. This decomposition locates a contribution to the intervention's gain, while the knockout results leave the natural failure pathway~unresolved.

\subsection{Do a Few Skill-Reading Heads Mediate the Loss?}
\label{sec:app-ha}

The proposed head account requires three links: a small set of reads must be necessary for expressing $A$, those reads must weaken in failing layouts, and replacing their outputs must restore composition. Testing only one link would leave the account underdetermined. The study uses nine digit-moving skill-seed units on each of 0.5B and 1.5B and three on 7B. Heads are ranked in \texttt{[A]x} by attention mass from digits, `=' and answer inputs to $A$'s block, excluding key 0, the attention sink.

\paragraph{Necessity holds for a minority of units.}
For necessity, we remove these reads in the top $k\in\{1,2,4,8,16,32\}$ heads and compare with layer-matched random sets. Here $k^*$ is the smallest set causing a presence drop of at least 50 points, exceeding the random 95th percentile by at least 30 points, with Holm-adjusted permutation $p<0.05$. The base-level criterion requires $k^*\leq16$ in at least two thirds of units.

The number of random sets must permit that decision. With $n$ sets and Holm correction over $m$ units, the attainable minimum is $m/(n+1)$. Before the study, we increased the proposed 100 sets on 0.5B and 1.5B and 30 sets on 7B to 400 and 150, respectively: the original floors, 0.089 and 0.097, could not have met 0.05. A qualifying set of at most sixteen heads exists in three of nine units on 0.5B, all \sk{SHIFT\_RIGHT} with four heads, and two of nine on 1.5B, both \sk{SHIFT\_RIGHT} with eight heads. None of three qualifies at 7B, where only \sk{SHIFT\_LEFT} in seed 2 reaches $k^*$, at thirty-two heads. The three qualifying 0.5B units have drops of 50.4 to 97.8 points against random 95th percentiles of 0.8 to 2.8 points, and the two qualifying 1.5B units have 100 and 84.2 points against 8.6 and 17.0 points (Figure~\ref{fig:fig_ha}(a)). Both bases fall short of six of nine units. The released code reproduces the per-unit drops, random-set percentiles and adjusted $p$ values.

\input{tables/fig-ha_float.tex}

\paragraph{Attention mass persists after the position gap.}
The collapse prediction requires top-$k^*$ attention mass, normalised to the single-skill condition, to be at most 0.5 in one call, after 20 random rows and after a gap of 20, but at least 0.8 after \sk{ID}, in two thirds of units. Units without $k^*$ fail this criterion by definition. The gap is especially diagnostic: it removes $A$'s output effect, yet attention ratios remain 0.970 to 0.993 on 0.5B, 0.974 to 1.004 on 1.5B and 0.972 on 7B. Reported one-call ratios are lower: 0.380 to 0.833, 0.172 to 0.307, and 0.618 and 0.387, respectively. Including sixteen-head sets, after the twenty-position gap presence is at most 6.0\% while attention ratios remain 0.898 to 1.004 (Figure~\ref{fig:fig_ha}(b)). Insertion-control presence uses its own random-row draw.

This contrast contradicts the proposed collapse-of-attention-mass signature. It does not show that the heads transmit the same information: attention mass measures where a head reads, whereas its output also depends on what is read and written.

\paragraph{Replacing the selected outputs does not restore composition.}
For sufficiency, we transplant these heads' outputs from \texttt{[A][ID]x}, whose answer is $A(x)$ rather than the composed answer. Success requires a composition gain of at least 10 points exceeding the random 95th percentile in two thirds of cells. Units without $k^*$ again fail by definition. Gains are at most 0.2, 0.8 and 0.4 points on the three bases, below 10 points (Figure~\ref{fig:fig_ha}(c)). For cells lacking $k^*$, descriptive top-sixteen gains are at most 1.0 point on 0.5B and 7B. Two 1.5B cells gain 16.4 points, and only one exceeds its random 95th percentile of 1.4 points, outside the specified qualifying-unit analysis.

All three base-level criteria therefore fail (Figure~\ref{fig:fig_ha}). The tested units do not support this small-head-set account, while leaving other head sets and information changes untested. Because every first skill here moves digits, the result does not cover \sk{IP}, \sk{MOD3} or \sk{MIRROR} as first skills.

\subsection{Do Broader Searches Reveal a Restorative Site at 7B?}
\label{sec:app-where-7b}

The initial 7B transplant study has only two cells, with gains of 6.4 and 0 points: seed 3 contributes no cell because its core pair retains $A$, and \sk{SHIFT\_RIGHT} in seed 1 achieves 100\% at length 8 but fails revised acceptance on held-out lengths. Two broader searches address coverage, each selected on 300 items and reported on 500 others.

\paragraph{A wider single-site search finds one success.}
It removes the layer-window restriction of the first search and permits `=' sites in later layers, over nine cells on up to four pairs per seed, eight with wrong-skill sources. Four cells include a constituent seed that works at length 8 but fails revised held-out acceptance. The specified criterion requires success in at least four of the eight, including a seed of \sk{SR}\sk{IP}. Only one cell succeeds: \sk{SWAP}\sk{IP} in seed 1, at the digit site of layer 17, reaches 32.2\% composition accuracy against 0\% for the wrong-skill source and a control-relative threshold of 11.8\%. Both constituents of that cell pass revised acceptance, and failed cells gain at most 2.6 points.

\paragraph{Replacing a layer band does not produce a general rescue.}
The band study asks whether a useful intervention requires sustained replacement across layers. It writes states from \texttt{[A][ID]x} into digit and `=' positions at every layer of a contiguous band, in five cells from seeds 2 and 3, where the skills work at length 8. The thirty-three candidate bands have widths 4, 8, 14 or 28 and starting layers spaced by two. Selection takes the narrowest band whose gain and margins over wrong-skill and norm-matched random-direction controls reach 10 points, or else the best candidate. Report-pool success requires the same gain and margins, and the specified localisation criterion requires three of five cells. Only \sk{SL}\sk{IP} in seed 2 succeeds: layers 4 to 7 raise accuracy from 0.6\% to 11.2\%, with margins of 11.2 and 10.2 points, against a control-relative threshold of 11.0\%. On the selection pool this cell loses its gain at fourteen layers, and the twenty-eight-layer replacement is near zero by construction. Other cells improve by at most 2.0 points, and replacement by a cell's own states leaves 100\% of outputs unchanged.

Across single sites, bands and selected head sets, no 7B test restores composition in more than one cell. Broader searches therefore do not identify a tested local intervention sufficient to explain the general digit failure. Distributed interactions, untested sites and limitations of the chosen sources and interventions remain competing explanations that these searches do not distinguish.

\subsection{What Does Restoring the Later Language Skill Identify?}
\label{sec:app-nb}

Language pairs tend to omit $B$, so their diagnosis targets the successor rather than assuming the digit result carries over. The question is whether prefix controls and restorative knockouts jointly distinguish three explanations: competition for execution, sensitivity to preceding material, and dependence on $B$'s own position. The study uses the eighteen different-field pairs, three skill seeds and 300 records on Qwen2.5-7B-Instruct, the primary base, and 0.5B-Instruct. A unit is a later skill and seed, eligible when presence is at least 80\% alone and falls by at least 30 points in one call.

\paragraph{Candidate explanations require joint signatures.}
Prefix controls place twenty tokens of neutral prose, ten \sk{ID} copies, random rows, \sk{INIT} rows, or the earlier skill's twenty-token terse description before soft $B$. Position-only controls shift $B$ by four or twenty positions. Attention interventions prevent $B$'s block from reading $A$, prevent all later positions from reading $A$, or do both, at every layer and head. Preventing later positions from reading the system message is the non-specific control.

Three combinations of prefix sensitivity and restorative knockouts were specified as signatures \emph{consistent with} candidate explanations. Intact $B$ after untrained material and shifts, together with restoration only by hiding $A$ from later positions, would support single-slot execution. Sensitivity to prose or rows, with restoration by hiding $A$ from $B$'s block, would support left-prefix layout fragility. Sensitivity to shifts without restoration after hiding $A$ from both query sets would support dependence on $B$'s own position. A base requires a majority of eligible units with a signature. Other combinations receive no mechanism interpretation.

\paragraph{Implementation checks bound the comparison.}
Validity checks require removal of the targeted skill when access to its own block is hidden, non-specific-control gain below 10 points, agreement of the full knockout with the twenty-position shift, and reproduction of stored outputs. In pilot runs, full texts were reproduced on only 93.7\% and 88.4\% of 0.5B and 7B items, below the planned 95\%. As allowed by the protocol, the reproduction check switched to item-level flags before the main runs. Passing this revised check establishes agreement of those flags, not exact reproduction of the full generated text.

A no-op knockout, which installs the hook without masking any edge, reproduces plain-run texts on only 89.1\% to 94.6\% of items, because knockout cells use unpadded per-length buckets while plain cells use left-padded batches. Pooled over three seeds (3,456 items per base), its metric offset from the plain run is at most 0.5 points (largest: $B$ present on 0.5B, $-0.5$ points, 95\% item-bootstrap interval $[-0.9,-0.1]$, the other three intervals include zero), and exact match agrees within 0.2 points. Every knockout gain is measured against the no-op knockout on the same path, so the offset does not enter the gains, and the check was not used as a gate.

Table~\ref{tab:app-nb} summarises both bases over eligible units, and the released code reproduces the per-unit results (a successor and seed averaged over three predecessor partners on 300 items). Prefix labels name the sensitivity pattern (soft only, functional text, prose/rows, rows only, position or mixed) and knockout labels the restoring intervention (either single knockout, only joint use, one side alone, neither or mixed). The sensitivity row recomputes the label without cells where more than 50\% of outputs omit the name or total field (collapse), and $A$ presence with $A$ hidden from later positions is the manipulation check.

\input{tables/app-nb-summary_float.tex}

\paragraph{On 0.5B, trained-prefix sensitivity has no majority knockout signature.}
Fourteen of eighteen units are eligible, and all required checks pass. Their mean $B$ presence is 100\% alone and 18.1\% in one call, but 72.5\% to 100\% after the insertion and shift controls. In nine of fourteen units, only trained soft $A$ lowers presence by at least 30 points while every other insertion and the twenty-position shift remains within 15 points of the single-skill result. Neutral insertions cause drops of at least 30 points in at most four of fourteen 0.5B units, whereas one call does so in all fourteen. No knockout pattern has a majority. The most common, in six units, is restoration by either single knockout.

On 0.5B, hiding $A$ from $B$'s block increases presence by 44.2 points, hiding it from later positions by 37.2 points, and hiding it from both by 81.9 points. The non-specific control changes it by $-0.1$ points. Item-bootstrap 95\% intervals of these gains and their 7B counterparts are narrower than 2 points. Either single intervention restores part of the successor effect. The full intervention almost restores it entirely, but reproduces the position-shift control by Remark~\ref{rem:shift}, so its gain is not independent pathway evidence.

\paragraph{On 7B, a large restoration gain does not isolate the cause.}
All eighteen units are eligible and pass the required checks. Presence falls from 99.5\% alone to 9.7\% in one call. Neutral prose and the earlier skill's text are already disruptive, yielding means 37.7\% and 20.5\% and drops of at least 30 points in eleven and fifteen units. No prefix-sensitivity class has a majority. Twelve of eighteen units restore $B$ only when $A$ is hidden from both query sets.

Mean 7B gains are 89.3 points for the full knockout, 33.0 for $B$'s block, 9.6 for later positions and $-0.8$ for the control. The full knockout again reproduces the shift control, whose presence is 98.8\%. Excluding eleven cells whose outputs usually omit the name or total field removes the knockout majority. The aggregate classification is therefore sensitive to malformed-output cells.

The combined signatures are unmet on both bases. This is more restrictive than asking whether any knockout improves $B$: improvement alone does not identify the route of natural interference or support the digit attention-disconnection account. The item-bootstrap intervals condition on observed units and seeds, omitting variation across them.

\paragraph{Layer bands locate effective interventions.}
A second stage asks where the selected knockout produces its gain, restricting it to each third of the layers. It uses the stronger single intervention, hiding $A$ from $B$'s block, on 0.5B, and the full knockout on 7B. The specified label \emph{localised} requires one third to recover at least half the full-depth gain while the others recover less than a quarter. The label \emph{distributed} means that no individual third reaches half. These labels describe the intervention response. The same words would require additional evidence if used to characterise the unmodified model's failure.

On 0.5B, band gains are 39.4, 7.2 and 0.2 points against 44.2 points over all layers and a half-gain threshold of 22.1 points. The first-third localisation label holds in eight of fourteen units. On 7B, gains are 43.7, 4.3 and 0.5 points against 89.3 points and a half-gain threshold of 44.7 points. The aggregate receives the distributed label, although its first third accounts for nearly all of the summed single-band gain. Individually, nine of eighteen units are localised in the first third and eight reach half in no third. No intervals were specified for band gains (Table~\ref{tab:app-nb}). The results locate where the intervention is effective, without identifying where the unmodified model's loss originates.

\subsection{Do Single-Skill Features Predict Interference?}
\label{sec:app-predict}

Predicting which pairs interfere from constituent features requires its own validation: a feature can correlate with an outcome without identifying its cause, and probe failure can reflect the probe rather than absence of information.

\paragraph{Held-out pair prediction is unsupported.}
The specified prediction study directly tests whether probe-derived single-skill features predict interfering pairs on Gemma-3-1B and Qwen2.5-7B. All eight held-out tests are evaluable and none is significant: $p$ ranges from 0.12 to 0.88 and $q$ from 0.63 to 0.88. An additional rank correlation over twenty-four pairs is $-0.16$ ($p=0.47$). These tests therefore do not validate the proposed use of single-skill features to predict pairwise interference.

\section{RQ3: Repair, Retention, and Transfer}
\label{sec:app-repair}

This appendix supports Section~\ref{sec:repair} (RQ3). The anchor is ordinary rehearsal, used to separate composition from retention rather than to introduce a new optimisation method. In Figure~\ref{fig:fig_retention}, lines connect the plain and anchored edit of one base, seed and edit side, not points of a training trajectory, and all smaller-base seeds and 7B seeds 2 and 3 are shown. In the tables below, bold marks the best value in a comparison group unless a caption says otherwise, yellow marks values quoted in the main text, lavender rows name panels, and teal and coral mark retained and lost accuracy or significant and nonsignificant tests.

\subsection{Composition Accuracy Does Not Measure~Retention}

Does a repair that raises composed accuracy keep the edited skill working alone? A predecessor edit updates only the twenty rows of $A$ and a successor edit only those of $B$, with the rest of the model frozen, from $N$ composed input-output examples (Section~\ref{sec:repair-ablation}). The anchored objective adds the same $N$ inputs under the edited skill alone with its own gold outputs, for $2N$ examples, and the unanchored objective uses only the composed examples. The anchor comparison therefore tests a standalone-task requirement together with the labelled examples that it adds.

\paragraph{Two outcomes require separate measurement.}
Within a pair, composition accuracy is the median over three training seeds, while standalone accuracy of the edited skill is their mean. Comparisons use one-sided Wilcoxon signed-rank tests over pairs~\citep{wilcoxon1945individual}, treating differences of at most 0.5 points as ties and applying Holm correction within each table~\citep{holm1979simple}. Two-sided SPoT comparisons are identified separately. The twenty-seven pairs share constituents, so these significance calculations are approximate.

Table~\ref{tab:app-ablation-seeds} shows that, across the nine smaller-base skill seeds and both edit sides at eight composed labels, medians for the edited skill alone range from 88.9\% to 99.0\% with anchoring and from 0.9\% to 22.0\% without it. At eight labels, anchored successor edits outperform plain successor edits in composed accuracy after Holm correction in only two of the nine smaller-base rows and in neither 7B seed. The anchored successor-retention entry for Gemma seed 2 records twenty-six pairs rather than twenty-seven. At the pair level, which the released code reproduces, the anchor advantage occurs in 661 of 701 comparisons, a descriptive count over shared skills rather than independent replications.

\input{tables/app-ablation-seeds_float.tex}

\paragraph{The 7B repetition preserves the distinction.}
The initial 7B seed has a failed \sk{MOD3} constituent, scoring 3.0\% at length 8. To check the result with usable constituents, we repeat the ablation and tuned comparisons at four and eight labels on seeds 2 and 3, where every skill reaches at least 89.0\% at that length. No edit-side or anchor effect on composition survives Holm correction there ($p=0.32$ to 1). At eight labels, standalone-retention medians range from 93.4\% to 97.4\% with anchoring and from 8.8\% to 23.0\% without it. A null composition difference therefore coexists with a large retention difference even when this constituent failure is excluded from the comparison.

\paragraph{The comparison does not match total supervision.}
There is no equivalence test for the edit-side or anchor null results: a non-significant composition difference cannot establish equal performance or equal retention. Anchored editing at $N$ labels sees $2N$ examples, and at four and eight labels both objectives take 50 optimiser steps. A post hoc, descriptive comparison therefore matches both, setting plain edits with eight composed labels against anchored edits with four composed labels and four anchors, on all 27 pairs, three seeds, both edit sides and four bases. Plain edits compose better in 492 of 648 pair, seed and side cells, tie in 136 and lose in 20, and win at least 14 of 27 pairs in each of the 24 rows. This count includes the failed 7B skill seed. Anchored edits with four labels stay near floor, with medians from 0\% to 6.3\%. At a fixed example budget, anchoring thus spends composed labels on the anchor, and its retention gain is measured only at eight labels. Retention is stored only at 1024 and 4096 labels and at the top budget of each run, so no retention pair is matched exactly. Plain edits given more examples than the eight-label anchored edit still lose the edited skill. On 7B skill seed 1, with its failed \sk{MOD3}, all four edits reach 93.0\% to 95.7\% composed accuracy at 64 labels and equal steps, yet plain $B$ and $A$ edits leave their skills at 14.8\% and 4.0\% (last row of Table~\ref{tab:app-ablation-seeds}).

\subsection{What Remains after Tuning Alternative Repairs?}
\label{sec:app-tuned}

The anchor comparison establishes a retention contrast between objectives but cannot rank repair methods, so we ask whether anchored editing outperforms repairs that leave the original skill rows unchanged. Our SPoT adaptation of soft-prompt transfer~\citep{vu2022spot} optimises a new forty-row pair prompt initialised from the two constituent blocks, and a connector optimises a twenty-row block inserted between the frozen constituents, separately for each pair. Both use $N$ composed examples, and all methods use Adam with the same number of optimiser steps at a given label budget. At inference, SPoT occupies the 40 positions of the original pair and the connector adds 20 positions, whereas SPoT trains 40 rows and each edit or connector trains 20 rows.

\paragraph{Tuning changes the apparent comparison.}
Each method retains the learning rate with the highest eight-label median over twenty-seven seed-1 pairs, from a grid evaluated at four and eight labels, and the grids differ across methods. The selected rate is fixed before evaluation at sixteen to sixty-four labels and on 7B seeds 2 and 3. Every selected rate is interior to its grid, and adjacent candidates differ by factors of 2.5 to 5 (Table~\ref{tab:app-lr}). SPoT is particularly sensitive. Early comparisons used the default 0.005 for both baselines, and apparent advantages over SPoT disappeared once its rate was tuned, so those default-rate comparisons support no superiority claim. Selection uses the compared pairs, without a held-out tuning set. Baselines have four candidate rates on most bases and anchored editing has three, so the search spaces are unequal, and interior selections do not resolve unequal search sizes or tuning on the observed pairs.

\input{tables/app-lr_float.tex}

\begin{table*}[!t]
\centering
\footnotesize
\setlength{\tabcolsep}{5pt}
\hlclip{\begin{tabular}[t]{@{}rccccc@{}}
\toprule
\hlheadrow \hlhd{Labels $N$} & \hlhd{Anch.} & \hlhd{SPoT} & \hlhd{Conn.} & \hlhd{$p_S$} & \hlhd{$H_C$} \\
\midrule
\hlgrouprow \multicolumn{6}{@{}l}{\emph{Qwen2.5-0.5B, seed 1}} \\
4 & 6.3 & \hlbest{6.7} & 1.3 & 0.26 & \hlpass 0.019 \\
8 & \hlbest{70.0} & 69.0 & 15.7 & 0.30 & \hlpass 0.0013 \\
16 & 87.7 & \hlbest{93.0} & 74.0 & 0.21 & \hlpass 0.0057 \\
32 & \hlbest{98.7} & \hlbest{98.7} & 95.3 & 0.31 & \hlpass 0.0057 \\
64 & \hlbest{100.0} & \hlbest{100.0} & \hlbest{100.0} & 0.42 & \hlfail 0.47 \\
\midrule
\hlgrouprow \multicolumn{6}{@{}l}{\emph{Qwen2.5-1.5B, seed 1}} \\
4 & \hlbest{0.3} & \hlbest{0.3} & 0.0 & 0.87 & \hlpass 0.0010 \\
8 & \hlbest{13.3} & 9.3 & 0.7 & 0.11 & \hlpass $2.8{\times}10^{-5}$ \\
16 & \hlbest{56.0} & 30.0 & 3.0 & 0.27 & \hlpass $1.9{\times}10^{-5}$ \\
32 & \hlbest{82.7} & 70.3 & 33.0 & 0.074 & \hlpass $4.5{\times}10^{-8}$ \\
64 & \hlbest{98.0} & 97.0 & 88.7 & 0.030 & \hlpass $1.9{\times}10^{-5}$ \\
\midrule
\hlgrouprow \multicolumn{6}{@{}l}{\emph{Gemma-3-1B, seed 1}} \\
4 & 3.3 & \hlbest{4.3} & 1.7 & 0.39 & \hlfail 0.069 \\
8 & 35.0 & \hlbest{38.7} & 11.0 & 0.89 & \hlpass 0.023 \\
16 & \hlbest{76.0} & 72.3 & 56.0 & 0.20 & \hlpass 0.0018 \\
32 & \hlbest{87.7} & \hlbest{87.7} & 87.3 & 0.044 & \hlfail 0.062 \\
64 & 98.7 & \hlbest{99.0} & 98.3 & 0.35 & \hlfail 0.30 \\
\bottomrule
\end{tabular}}\hspace{20pt}
\hlclip{\begin{tabular}[t]{@{}rccccc@{}}
\toprule
\hlheadrow \hlhd{Labels $N$} & \hlhd{Anch.} & \hlhd{SPoT} & \hlhd{Conn.} & \hlhd{$p_S$} & \hlhd{$H_C$} \\
\midrule
\hlgrouprow \multicolumn{6}{@{}l}{\emph{Qwen2.5-7B, seed 2}} \\
4 & 0.7 & \hlbest{1.7} & 0.3 & 0.42 & \hlfail 0.22 \\
8 & 9.3 & \hlkey \hlbest{15.0} & 8.0 & 0.47 & \hlfail 0.065 \\
16 & 37.0 & \hlbest{43.7} & 39.0 & 0.99 & \hlfail 0.63 \\
32 & 67.0 & 57.3 & \hlbest{68.7} & 0.0019 & \hlfail 0.20 \\
64 & \hlbest{94.7} & 94.0 & 90.7 & 0.0020 & \hlfail 0.073 \\
\midrule
\hlgrouprow \multicolumn{6}{@{}l}{\emph{Qwen2.5-7B, seed 3}} \\
4 & \hlbest{0.7} & \hlbest{0.7} & 0.3 & 0.044 & \hlfail 0.099 \\
8 & 7.0 & \hlkey \hlbest{15.3} & 1.3 & 0.069 & \hlpass $4.8{\times}10^{-4}$ \\
16 & 25.0 & \hlbest{29.7} & 4.3 & 0.21 & \hlpass 0.012 \\
32 & \hlbest{68.7} & 66.0 & 10.0 & 0.88 & \hlpass 0.0045 \\
64 & \hlbest{92.0} & 90.7 & 67.7 & 0.75 & \hlpass $2.1{\times}10^{-4}$ \\
\midrule
\hlgrouprow \multicolumn{6}{@{}l}{\emph{Qwen2.5-7B, seed 1$^\dagger$}} \\
4 & 0.0 & \hlbest{0.3} & \hlbest{0.3} & 0.81 & \hlfail 0.068 \\
8 & 4.7 & \hlbest{9.3} & 3.7 & 0.55 & \hlfail 0.068 \\
16 & \hlbest{30.3} & 25.3 & 15.7 & 0.65 & \hlfail 0.18 \\
32 & 65.0 & \hlbest{67.7} & 52.7 & 0.60 & \hlpass 0.049 \\
64 & \hlbest{95.3} & 88.7 & 92.3 & 0.071 & \hlfail 0.47 \\
\bottomrule
\end{tabular}}

\caption{\textbf{No repair method ranks consistently first.} Medians (\%) over 27 pairs for anchored successor editing (Anch.), SPoT and the connector, $p_S$ unadjusted two-sided versus SPoT, $H_C$ Holm one-sided versus the connector.}
\label{tab:tuned}
\end{table*}

\paragraph{The tuned results do not yield a consistent ranking.}
Table~\ref{tab:tuned} reports every budget at the selected rates, with 7B seed 1 ($\dagger$, failed \sk{MOD3}) after seeds 2 and 3. At eight labels on seed 1, anchored editing is not separated from SPoT on 0.5B, 1.5B or Gemma (unadjusted two-sided $p=0.30$, 0.11 and 0.89). On 7B seeds 2 and 3, SPoT has higher median composition accuracy, 15.0\% versus 9.3\% and 15.3\% versus 7.0\%, without significant separation in those comparisons ($p=0.47$ and 0.069). No 7B SPoT comparison at four or eight labels survives Holm correction.

At eight labels, anchored editing exceeds the per-pair connector after Holm correction in ten of the 12 base and seed rows, all except 7B seeds 1 and 2 ($p=0.068$ and 0.065), with twice as many examples. Isolated cells favour anchored editing against SPoT on 1.5B seed 2 at eight labels, 1.5B at sixty-four labels and 7B seed 2 at thirty-two and sixty-four labels, with Holm-adjusted one-sided $p=0.035$, 0.044 and 0.0057 in both 7B cells. Table~\ref{tab:tuned} lists two-sided values instead and omits 1.5B seed 2. Others favour SPoT at four labels on seed 3 of Gemma and 7B (unadjusted two-sided $p=0.037$ and 0.044), although both 7B medians equal 0.7\%. On 0.5B at eight labels, SPoT brings ten pairs to at least 90\%, compared with six for anchored editing. When examples and steps are both matched, at eight examples and 50 steps, tuned SPoT with eight composed labels exceeds anchored editing with four on 16 to 24 of 27 pairs in each of the 12 base and seed rows, 250 of 324 cells in all (post hoc). The tuned connector does so in 167 of 324 cells, but not on most 1.5B pairs.

Equal steps do not remove anchoring's additional labelled examples or the differences in parameterisation. These isolated advantages therefore establish neither consistent superiority nor equivalence. Methods can obtain similar composition scores while achieving retention through different constraints, including leaving the constituent rows unchanged, so repair accuracy and retention should be reported separately.

\subsection{Does the Retention Contrast Extend to Language Skills?}

Does the separation between composition and retention also appear when the missing effect is a record-editing successor? The language study evaluates successor repair on the twenty interfering Qwen2.5-0.5B-Instruct pairs, with successor editing, anchored successor editing, a pair prompt and a connector, all at the untuned default rate 0.005. Equiv. and Non-inf. count pairs whose 90\% paired item-bootstrap interval of one-call minus two-call accuracy (300 items, 10{,}000 resamples, margins fixed before the runs) lies within $\pm 10$ points or has its lower bound above $-2$ points, respectively. Two-call accuracy is the higher of its zero-shot value and its value with the current skill rows, and a pair counts at $N$ once its criterion holds on at least two of three training seeds at some budget up to $N$. The final column is the Holm-adjusted one-sided $p$ that anchored editing exceeds the corresponding mode.

Table~\ref{tab:app-nl-repair} shows that composition improves under all four modes. At sixteen labels, the 90\% interval of ten to fourteen pairs lies within 10 points of two-call accuracy. Anchored editing exceeds no alternative after Holm correction, and the closest comparison is SPoT at sixteen labels ($p=0.051$). Retention separates these results across the sixteen-to-1024-label budgets. Unanchored editing lowers $B$ alone by a median 86.0 to 100 points, whereas anchored editing lowers it by 0 to 0.3 points. SPoT and the connector retain $B$'s original rows and hence do not change its standalone execution under the masked-output protocol.

\input{tables/app-nl-repair_float.tex}

The same qualitative contrast therefore appears in this language setting: improving the pair need not preserve the constituent. Untuned baselines and coverage of only one base and edit side limit the result to that observation, without establishing a ranking of repair methods.

\subsection{Can More Labels Repair the Pair without Erasing Its Skills?}

The low-label comparisons leave open whether high composition accuracy is compatible with retention across most pairs. A broader-budget study applies anchored successor editing to all twenty-seven seed-1 pairs of the three smaller digit bases at rate 0.005. Starting from a zero-shot median of 0\%, it reaches 90\% composition accuracy by sixty-four labels on 27, 20 and 25 pairs of 0.5B, 1.5B and Gemma, respectively. At 1024 labels, it reaches 95\% on twenty-six of twenty-seven pairs on each base. At the largest evaluated budgets, the maximum drop of either constituent is at most 5 points for every pair. The tested objective can therefore combine accurate execution of the repaired pair with constituent retention at larger budgets.

This is a claim about the repaired pair and its constituents. Other pairs ending in the edited successor can change: the worst per-pair mean change at eight labels is 12.8 to 19.8 points on the smaller bases. To assess reuse across a library, we must separately evaluate new partners and existing~\mbox{contexts}.

\subsection{Does a Repaired Skill Transfer to Other Partners?}
\label{sec:app-transfer}

\input{tables/app-transfer_float.tex}

\begin{figure*}[t]
\centering
\includegraphics[width=\textwidth]{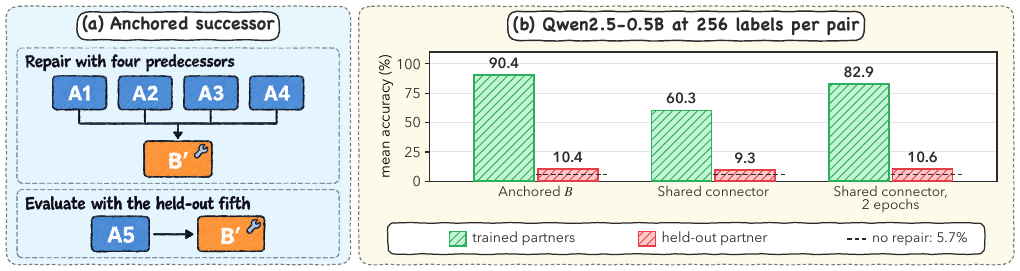}
\caption{\textbf{Successful repair on training partners transfers weakly to an unseen predecessor.} Bars are fold means at 256 composed labels per training pair, and the 5.7\% no-repair baseline applies only to the held-out partner.}
\label{fig:partner-transfer}
\end{figure*}

Standalone retention checks the original skill, but does not show that the edit becomes compatible with other partners. Table~\ref{tab:app-transfer} separates three evaluations: an unseen learned partner, other learned partners after a single-pair edit, and pretrained successors after a predecessor edit. On Qwen2.5-0.5B with skill seed 1, each of the twenty-seven folds holds out one non-degenerate directed pair and trains its anchored successor on the successor's other non-degenerate predecessors, four for \sk{IP}, \sk{SWAP} and \sk{MIRROR} and three for \sk{SR}, \sk{SL} and \sk{MOD3}, so no degenerate pair is trained or tested. Each fold reports its median over three training seeds on 300 evaluation inputs, means are taken over folds, and a fold or pair gains when its median exceeds its no-training accuracy by at least 10 points. Figure~\ref{fig:partner-transfer} draws a fold with four training predecessors, in which the same repaired successor meets its training predecessors and the unseen one, and the 256-label means for anchored successor editing and shared connectors.

\paragraph{Training on several partners gives little held-out gain.}
At 256 composed labels per training pair, mean accuracy is 90.4\% on trained partners and 10.4\% on the held-out partner. The held-out mean without repair is 5.7\%, giving a gain of 4.7 points. Only four of twenty-seven folds improve by at least 10 points. A shared connector shows similarly limited transfer: held-out accuracy is 9.3\% at this budget and 10.6\% with twice the epochs, with two and three folds gaining at least 10 points, respectively. It uses the edits' rate of 0.005 rather than the rate of 0.02 tuned for the per-pair connector, and fits its training pairs only partly. These results do not support broad transfer of the repaired successor to an unseen predecessor.

\paragraph{A connector shared by all training pairs.}
A post hoc check asks whether sharing one connector more widely helps. A single 20-row connector is trained on the composed labels of 18 non-degenerate directed pairs and evaluated on the other nine, in three folds that hold out each of the 27 pairs once, with the initialisation, step rule and evaluation of the per-successor connector (rows marked all pairs in Table~\ref{tab:app-transfer}). At 256 labels per training pair, the held-out mean is 7.7\% against 5.7\% without repair, and two of 27 pairs gain at least 10 points. With twice the epochs, the connector reaches 72.4\% on its training pairs and 9.0\% on held-out pairs, with three pairs gaining. At the rate of 0.02 tuned for the per-pair connector, the held-out mean is 8.5\%. A connector shared by all training pairs thus fits them only partly and transfers no better than one shared by the predecessors of a single successor.

After single-pair editing, median changes on the edited skill's other learned partners are also near zero across bases. Such medians do not imply that every other pair is unaffected, as the changes above demonstrate. Under a strategy that stores a separate twenty-row repair for every ordered pair in a $K$-skill library, the number of copies can reach $K(K-1)$. The transfer measurements provide no basis for assuming those copies can be replaced by one generally compatible edit. No matched training-partner baseline is supplied, and anchored editing adds own-task rehearsal, so the methods are not matched in total training examples. The folds share skill structure, and no new uncertainty estimates are inferred from these aggregates.

\paragraph{Retaining the standalone task does not preserve every old context.}
\remarkpar{rem:causal}With the base frozen and novel output IDs masked, editing only $B$'s input rows cannot change execution in a prompt that omits $B$. Editing $A$ changes the module used in every context containing $A$. Under causal attention, suffixes also cannot change hidden states at $A$'s own positions.\par

A predecessor edit also changes contexts in which it is followed by pretrained operations. At eight labels on seed 1, anchored predecessor editing reduces accuracy with a following \sk{ID} by a median 9.1 to 31.7 points across bases. The 7B seed in this comparison contains the failed \sk{MOD3}. At 4096 labels, evaluated on fewer pairs but every skill seed, accuracy with a following \sk{ADD} changes by $-21.3$ to $-95.2$ points under anchoring and by $-71.1$ to $-95.2$ points without it. The budgets use different pair sets and cannot be interpreted as a learning curve. These are additional context-retention measurements, not held-out-partner training experiments. Standalone replay therefore cannot certify all earlier uses of an updated module.

A successor edit leaves those $A$-only contexts unchanged: all 2416 observed cells change by 0 points. This is a structural control (Remark~\ref{rem:causal}). It does not demonstrate transfer. No experiment trains a predecessor edit on several successors and evaluates it on a held-out successor. That corresponding direction remains untested.

\input{tables/app-cost_float.tex}

\subsection{When Can a Pair-Specific Repair Amortise Its Cost?}
\label{sec:app-cost}

The limited transfer makes the serving question depend on how often each repaired pair is used. Two calls require another model invocation but no repair training. A learned one-call repair pays an initial cost that must be recovered through repeated requests, so the comparison must consider the achieved accuracy as well as latency.

\paragraph{The extra invocation has a measured latency cost.}
Dedicated timing on one H100 measures four interfering Qwen2.5-0.5B pairs. For three, two calls take 2.00 to 2.01 times one-call request latency at batch size 1 and 1.93 to 1.99 times at batch size 32. The fourth, \sk{SWAP}\sk{SR}, has an unusually short incorrect one-call output that terminates early. Across the four bases and three skill seeds, two calls take between 1.61 and 2.06 times as long in batched evaluation of unrepaired prompts (per-base means 1.67 to 1.95, Table~\ref{tab:app-cost}). The token budgets differ less. One call places both 20-row blocks in a prompt of 49 tokens on every digit base, against 58 tokens summed over two calls (Table~\ref{tab:app-cost}). For the language pairs the means are about 101 and 163 tokens on all three Instruct sizes.

\paragraph{Break-even depends on batching and repair quality.}
\looseness=-1 On 0.5B, anchored editing at sixty-four labels matches two-call accuracy at 100\% and amortises its training after approximately 283 batch-1 requests, compared with 320 for SPoT (Table~\ref{tab:app-cost}). Under batched serving, the anchored break-even count is approximately 11,115. At eight labels, the estimated break-even count is 112, but accuracy is only 55.0\% rather than 100\%. The smaller count therefore does not offer equal service quality.

Training wall times include an evaluation and shared-GPU contention, and are reported as upper-bound measurements. They do not increase monotonically with model size, so amortisation comparisons are meaningful within a base rather than across bases. Training and storage costs recur for the pairs served by a library, and tuned SPoT shares the one-call saving. These measurements therefore support a conditional benefit of frequently reused pair repairs, without establishing that constituent editing is the most economical way to obtain it.

\clearpage
\onecolumn
\section{Implications for Agent Skill Libraries}
\label{sec:app-agent-implications}
Agent libraries mostly use text or code skills (Section~\ref{sec:related}), so our parametric results motivate tests for them rather than establish their outcomes. This appendix extends Section~\ref{sec:agent-implications} with those tests, and Figure~\ref{fig:agent-library-scope} separates the evaluated interface from the prospective decisions.\par\vspace{0pt}
{\centering
\includegraphics[width=0.58\textwidth]{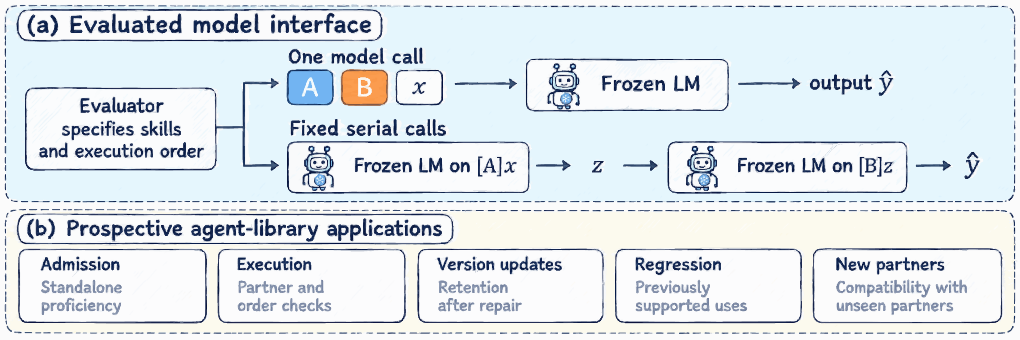}\par\vspace{-4pt}}
\captionof{figure}{\textbf{Model-level evidence motivates prospective agent-library evaluations.} Panel (a) is the evaluated interface and panel (b) lists prospective applications, an application framework rather than an implemented or evaluated controller.}
\label{fig:agent-library-scope}
\vspace{-8pt}

\begin{multicols}{2}\csname nolinenumbers\endcsname %
Skill libraries already support exploration in embodied environments and automation of application tasks: Voyager stores executable behaviours, while AXIS acquires and reuses validated application skills~\citep{wang2024voyager,lu2025axis}. We instead measure independently learned soft input embeddings in a frozen backbone, with skill identities and order supplied by the evaluator and serial execution adding a model call and an explicit intermediate output. The prospective layer covers admission evidence, execution choices, version maintenance and tests on new partners for agents adopting this interface, and applying it to textual or executable skill libraries would require separate measurement.

\paragraph{Admission and execution.}
\label{sec:app-agent-admission}
Acceptance on the original task establishes less than acceptance for joint use, because constituent proficiency does not establish the ordered combinations an entry adds (Section~\ref{sec:revisit-interfere}). A controller may select both operations correctly while their joint invocation fails, and neutral-insertion robustness does not substitute for partner-specific evidence (Section~\ref{sec:diag-compat}). Our pairwise results motivate choosing which likely combinations to test as libraries grow, but do not validate a sampling strategy or establish reliability of longer compositions. Separate calls execute the tested pairs under this scaffold (Section~\ref{sec:revisit-interfere}), but they add computation and intermediate text, so the control does not isolate which addition explains the gain or establish a universally preferable execution policy. Procedures and examples help most 7B different-field record edits while order-dependent pairs still interfere (Section~\ref{sec:diag-carrier}), so representation and execution are best evaluated together, with their information differences explicit and with accuracy, latency and repair cost judged jointly. Amortisation depends on reuse frequency, batching and achieved accuracy (Appendix~\ref{sec:app-cost}).

\paragraph{Replacement and updates.}
\label{sec:app-agent-versions}
A repaired entry can satisfy one new combination while losing its original task, and standalone rehearsal can leave other supported contexts degraded (Section~\ref{sec:repair-ablation} and Appendix~\ref{sec:app-transfer}), so an update changes the entry's tested uses, not only its parameters. With novel output IDs masked, editing only a skill's input rows leaves prompts without those rows unchanged, but this structural guarantee does not extend to prompts containing the edited skill. A separate pair prompt keeps the constituent rows but adds a composition-specific object whose cost and reuse must also be evaluated (Appendix~\ref{sec:app-tuned}). These maintenance consequences warrant examination without establishing a preferred versioning policy or a superior repair method.

\paragraph{Planning and agent-level tests.}
\label{sec:app-agent-planning}
Successful repair on training partners does not establish reuse in a new combination: a repaired successor transfers weakly to a new predecessor (Appendix~\ref{sec:app-transfer}), so a controller would need to distinguish observed from proposed combinations and assess uncertainty about the latter explicitly. An agent-level evaluation, beyond the present study, would hold the goals and skill pool fixed, separate incorrect skill selection from failed execution of a correct selection, and report task success with invocation and repair costs. It would evaluate new-partner combinations before they receive repair supervision and recheck updates on previously supported uses, while longer sequences, richer language tasks and feedback-driven execution would test further forms of reuse. Such experiments could establish whether model-level diagnostics improve downstream decisions, a benefit that the current controlled results alone cannot~demonstrate.
\end{multicols}\csname linenumbers\endcsname

\end{document}